\documentclass{article}

\usepackage{microtype}
\usepackage{graphicx}
\usepackage{booktabs}
\usepackage{siunitx} 
\usepackage{enumitem}
\usepackage{tikz}
\usepackage{url}     
\usepackage{amsfonts}
\usepackage{nicefrac}
\usepackage{microtype}
\usepackage{subcaption}

\usepackage{amsmath, amsthm}
\usepackage{graphicx}

\usepackage{hyperref}

\usepackage[accepted]{icml2019}

\makeatletter
\usepackage{footmisc}
\DefineFNsymbols{mySymbols}{{\ensuremath\star}{\ensuremath\ddagger}\S\P
  *{**}{\ensuremath{\dagger\dagger}}{\ensuremath{\ddagger\ddagger}}}
\setfnsymbol{mySymbols}
\usepackage{xcolor}
\usepackage{hyperref}
\hypersetup{
    colorlinks,
    anchorcolor={blue!50!black},
    linkcolor={blue!50!black},
    citecolor={blue!50!black},
    urlcolor={blue}}

\newcommand{\R}{\mathbb{R}}
\newcommand{\E}{\mathbb{E}}

\newcommand{\celebahq}{\textsc{celeba-hq-128}}
\newcommand{\lsun}{\textsc{lsun-bedroom}}

\DeclareMathOperator{\Tr}{Tr}

\newcommand{\pdata}{P}
\newcommand{\pmodel}{Q}
\newcommand{\ld}{\mathcal{L}_\textsc{d}}
\renewcommand{\lg}{\mathcal{L}_\textsc{G}}

\icmltitlerunning{A Large-Scale Study on Regularization and Normalization in GANs}

\begin{document}
\twocolumn[
\icmltitle{A Large-Scale Study on Regularization and Normalization in GANs}
\icmlsetsymbol{equal}{*}

\begin{icmlauthorlist}
\icmlauthor{Karol Kurach}{equal,brain}
\icmlauthor{Mario Lucic}{equal,brain}
\icmlauthor{Xiaohua Zhai}{brain}
\icmlauthor{Marcin Michalski}{brain}
\icmlauthor{Sylvain Gelly}{brain}
\end{icmlauthorlist}

\icmlaffiliation{brain}{Google Research, Brain Team}
\icmlcorrespondingauthor{Karol Kurach}{kkurach@google.com}
\icmlcorrespondingauthor{Mario Lucic}{lucic@google.com}

\icmlkeywords{GANs, regularization, normalization, large-scale study}
\vskip 0.3in
]
\printAffiliationsAndNotice{\icmlEqualContribution}

\begin{abstract} Generative adversarial networks (GANs) are a class of deep
  generative models which aim to learn a target distribution in an unsupervised
  fashion. While they were successfully applied to many problems, training a GAN
  is a notoriously challenging task and requires a significant number of
  hyperparameter tuning, neural architecture engineering, and a non-trivial
  amount of ``tricks". The success in many practical applications coupled with
  the lack of a measure to quantify the failure modes of GANs
  resulted in a plethora of proposed losses, regularization and normalization schemes,
  as well as neural architectures. In this work we take a sober view of the current
  state of GANs from a practical perspective. We discuss and evaluate common pitfalls and reproducibility issues, open-source our code on Github, and provide pre-trained models on TensorFlow Hub.
\end{abstract}

\section{Introduction}\label{sec:introduction}
Deep generative models are a powerful class of (mostly) unsupervised machine learning models. These models were recently applied to great effect in a variety of applications, including image generation, learned compression, and domain adaptation~\citep{brock2018large,menick2018generating,karras2018style,lucic2019high,isola2017image,tschannen2018deep}.

Generative adversarial networks (GANs)~\citep{goodfellow2014generative} are one of the main approaches to learning such models in a fully unsupervised fashion. The GAN framework can be viewed as a two-player game where the first player, the \emph{generator}, is learning to transform some simple input distribution to a complex high-dimensional distribution (e.g. over natural images), such that the second player, the \emph{discriminator}, cannot tell whether the samples were drawn from the true distribution or were synthesized by the generator. The solution to the classic minimax formulation~\citep{goodfellow2014generative} is the Nash equilibrium where neither player can improve unilaterally. As the generator and discriminator are usually parameterized as deep neural networks, this minimax problem is notoriously hard to solve.

In practice, the training is performed using stochastic gradient-based optimization methods. Apart from inheriting the optimization challenges associated with training deep neural networks, GAN training is also sensitive to the choice of the loss function optimized by each player, neural network architectures, and the specifics of regularization and normalization schemes applied. This has resulted in a flurry of research focused on addressing these challenges ~\citep{goodfellow2014generative,salimans2016improved,miyato2018spectral,gulrajani2017improved,arjovsky2017wasserstein,mao2016least}. 

\textbf{Our Contributions}\quad In this work we provide a thorough empirical analysis of these competing approaches, and help the researchers and practitioners navigate this space. We first define the GAN landscape -- the set of loss functions, normalization and regularization schemes, and the most commonly used architectures. We explore this search space on several modern large-scale datasets by means of hyperparameter optimization, considering both ``good'' sets of hyperparameters reported in the literature, as well as those obtained by sequential Bayesian optimization. 

We first decompose the effect of various normalization and regularization schemes. We show that both gradient penalty~\citep{gulrajani2017improved} as well as spectral normalization~\citep{miyato2018spectral} are useful in the
context of high-capacity architectures. Then, by analyzing the impact of the loss
function, we conclude that the non-saturating loss~\cite{goodfellow2014generative} is sufficiently stable across datasets and hyperparameters. Finally, show that similar conclusions hold for both popular types of neural architectures used in state-of-the-art models.
We then discuss some common pitfalls, reproducibility issues, and practical considerations. We provide reference implementations, including training and evaluation code on
\href{http://www.github.com/google/compare_gan}{Github}\footnote{\url{www.github.com/google/compare_gan}},
and provide pre-trained models on \href{http://www.tensorflow.org/hub}{TensorFlow Hub}\footnote{\url{www.tensorflow.org/hub}}.

\section{The GAN Landscape}
The main design choices in GANs are the loss function, regularization and/or normalization approaches, and the neural architectures. At this point GANs are extremely sensitive to these design choices. This fact coupled with optimization issues and hyperparameter sensitivity makes GANs hard to apply to new datasets. Here we detail the main design choices which are investigated in this work.

\subsection{Loss Functions} Let $P$ denote the target (true) distribution
and $Q$ the model distribution. \citet{goodfellow2014generative} suggest two loss functions: the minimax GAN and the non-saturating (NS) GAN. In the former the discriminator minimizes the negative log-likelihood for the binary classification task. In the latter the generator maximizes the probability of generated samples being real. In this work we consider the non-saturating loss as it is known to outperform the minimax variant empirically. The corresponding discriminator and generator loss functions are
\begin{align*}
\ld &= -\E_{x \sim \pdata}[\log(D(x))] - \E_{\hat{x} \sim \pmodel}[\log(1 - D(\hat{x}))],\\
\lg &= -\E_{\hat{x} \sim \pmodel}[\log(D(\hat{x}))],
\end{align*}
where $D(x)$ denotes the probability of $x$ being sampled from $\pdata$.
In Wasserstein GAN (WGAN)~\citep{arjovsky2017wasserstein} the authors propose to consider the Wasserstein distance instead of the Jensen-Shannon (JS) divergence. The corresponding loss functions are
\begin{align*}
    \ld &= -\E_{x \sim \pdata}[D(x)] + \E_{\hat{x} \sim \pmodel}[D(\hat{x})],\\
    \lg &= - \E_{\hat{x} \sim \pmodel}[D(\hat{x})],
\end{align*}
where the discriminator output $D(x) \in \R$ and $D$ is required to be 1-Lipschitz. Under the optimal discriminator, minimizing the proposed loss function with respect to the generator minimizes the Wasserstein distance between $P$ and $Q$. A key challenge is ensure the Lipschitzness of $D$. Finally, we consider the least-squares loss (LS)
which corresponds to minimizing the Pearson $\chi^2$ divergence between $P$ and
$Q$~\citep{mao2016least}. The corresponding loss functions are
\begin{align*}
\ld &= -\E_{x \sim \pdata}[(D(x) - 1)^2] + \E_{\hat{x} \sim \pmodel}[D(\hat{x})^2],\\
  \lg &= -\E_{\hat{x} \sim \pmodel}[(D(\hat{x}) - 1)^2],
\end{align*} 
where $D(x) \in \R$ is the output of the discriminator. Intuitively, this loss smooth loss function saturates slower than the cross-entropy loss.

\subsection{Regularization and Normalization}

\looseness=-1\textbf{Gradient Norm Penalty}\quad The idea is to regularize $D$ by constraining the norms of its gradients (e.g. $L_2$). In the context of Wasserstein GANs and optimal transport this regularizer arises naturally and the gradient norm is evaluated on the points from the \emph{optimal coupling} between samples from $\pdata$ and $\pmodel$ (GP)~\citep{gulrajani2017improved}. Computing this coupling during GAN training is computationally intensive, and a linear interpolation between these samples is used instead. The gradient norm can also be penalized close to the data manifold which encourages the discriminator to be piece-wise linear in that region (Dragan)~\citep{kodali2017dragan}. A drawback of gradient penalty (GP) regularization scheme is that it can depend on the model distribution $Q$ which changes during training. For Dragan it is unclear to which extent the Gaussian assumption for the manifold holds. In both cases, computing the gradient norms implies a non-trivial running time overhead.

Notwithstanding these natural interpretations for specific losses, one may also consider the gradient norm penalty as a classic regularizer for the complexity of the discriminator~\citep{fedus2018many}. To this end we also investigate the impact of a $L_2$ regularization on $D$ which is ubiquitous in supervised learning.

\textbf{Discriminator Normalization}\quad
Normalizing the discriminator can be useful from both the optimization
perspective (more efficient gradient flow, more stable optimization), as well
as from the representation perspective -- the representation richness of the
layers in a neural network depends on the spectral structure of the
corresponding weight matrices~\citep{miyato2018spectral}.

\looseness=-1From the optimization point of view, several normalization techniques
commonly applied to deep neural network training have been applied to
GANs, namely batch normalization (BN)~\citep{ioffe2015batch} and layer
normalization (LN)~\citep{ba2016layer}. The former was explored
in~\citet{denton2015deep} and further popularized
by~\citet{radford2015unsupervised}, while the latter was investigated
in~\citet{gulrajani2017improved}. These techniques are used to
normalize the activations, either across the batch (BN), or across
features (LN), both of which were observed to improve the empirical
performance.

From the representation point of view, one may consider the neural network as
a composition of (possibly non-linear) mappings and analyze their spectral
properties. In particular, for the discriminator to be a bounded operator
it suffices to control the operator norm of each mapping. This approach is followed
in~\citet{miyato2018spectral} where the authors suggest dividing each weight
matrix, including the matrices representing convolutional kernels, by their
spectral norm. It is argued that spectral normalization results in discriminators of higher rank with respect to the competing approaches.

\subsection{Generator and Discriminator Architecture}
We explore two classes of architectures in this study: deep convolutional
generative adversarial networks (DCGAN)~\citep{radford2015unsupervised} and residual networks (ResNet)~\citep{he2016deep}, both of which are ubiquitous in GAN research. Recently,~\citet{miyato2018spectral} defined a variation of DCGAN, so called~\emph{SNDCGAN}. Apart from minor updates (cf. Section~\ref{sec:pitfalls}) the main difference to DCGAN is the use of an eight-layer discriminator network. The details of both networks are summarized in Table~\ref{tab:sndcgan_architecture}. The other architecture, \emph{ResNet19}, is an architecture with five ResNet blocks in the generator and six ResNet blocks in the discriminator, that can operate on $128\times128$ images. We follow the ResNet setup from~\citet{miyato2018spectral}, with the small difference that we simplified the design of the discriminator.

The architecture details are summarized in Table~\ref{tab:resnet19_discriminator} and Table~\ref{tab:resnet19_generator}. With this setup we were able to reproduce the results in~\citet{miyato2018spectral}. An ablation study on various ResNet modifications is available in the Appendix.

\subsection{Evaluation Metrics}\label{sec:metrics} We focus on several recently proposed metrics well suited to the image domain. For an in-depth overview of quantitative metrics we refer the reader to~\citet{borji2018pros}.

\textbf{Inception Score (IS)}\quad Proposed by \citet{salimans2016improved}, the IS
offers a way to quantitatively evaluate the quality of generated samples.
Intuitively, the conditional label distribution of samples containing meaningful
objects should have low entropy, and the variability of the samples should be
high. which can be expressed as $\texttt{IS} = \exp(\E_{x \sim Q}[d_{KL}(p(y
\mid x), p(y))])$. The authors found that this score is well-correlated with
scores from human annotators. Drawbacks include insensitivity to the prior distribution over labels and not being a proper \emph{distance}.

\textbf{Fr\'echet Inception Distance (FID)}\quad In this approach proposed by~\citet{heusel2017gans} samples from $P$ and $Q$ are first embedded into a feature
space (a specific layer of InceptionNet). Then, assuming that the embedded data
follows a multivariate Gaussian distribution, the mean and covariance are
estimated. Finally, the Fr\'echet distance between these two Gaussians is
computed, i.e.
\[
  \texttt{FID} = ||\mu_x - \mu_y||_2^2 + \Tr(\Sigma_x + \Sigma_y -
  2(\Sigma_x\Sigma_y)^\frac12),
\]
where $(\mu_x, \Sigma_x)$, and $(\mu_y, \Sigma_y)$ are the mean and covariance
of the embedded samples from $P$ and $Q$, respectively. The authors argue that
FID is consistent with human judgment and more robust to noise than IS.
Furthermore, the score is sensitive to the visual quality of generated samples
-- introducing noise or artifacts in the generated samples will reduce the FID.
In contrast to IS, FID can detect intra-class mode dropping -- a model that
generates only one image per class will have a good IS, but a bad FID~\citep{lucic2017gans}. 

\textbf{Kernel Inception Distance (KID)}\quad \citet{binkowski2018demystifying} argue that FID has no unbiased estimator and suggest KID as an unbiased alternative. In Appendix~\ref{sec:fid_kid_comparison} we empirically compare KID to FID and observe that both metrics are very strongly correlated (Spearman rank-order correlation coefficient of $0.994$ for $\lsun$ and $0.995$ for $\celebahq$ datasets). As a result we focus on FID as it is likely to result in the same ranking.

\subsection{Datasets}
We consider three datasets, namely \textsc{cifar10}, \celebahq{}, and \lsun{}. The $\lsun$ dataset contains slightly more than 3 million images~\citep{yu15lsun}.\footnote{The images are preprocessed to $128\times\!128\times3$ using TensorFlow resize\_image\_with\_crop\_or\_pad.} We randomly partition the images into a train and test set whereby we use 30588 images as the test set. Secondly, we use the \textsc{celeba-hq} dataset of $30$K images~\citep{karras2017progressive}. We use the $128\!\times\!128\!\times\!3$ version obtained by running the code provided by the authors.\footnote{ \url{github.com/tkarras/progressive\_growing\_of\_gans}} We use $3$K examples as the test set and the remaining examples as the training set. Finally, we also include the \textsc{cifar10} dataset which contains $70$K images ($32\!\times\!32\!\times\!3$), partitioned into $60$K training instances and $10$K testing instances. The baseline FID scores are 12.6 for $\celebahq$, 3.8 for $\lsun$, and 5.19 for \textsc{cifar10}. Details on FID computation are presented in Section~\ref{sec:pitfalls}.

\subsection{Exploring the GAN Landscape}

The search space for GANs is prohibitively large: exploring all combinations
of all losses, normalization and regularization schemes, and architectures is
outside of the practical realm. Instead, in this study we analyze several
slices of this search space for each dataset. In particular, to ensure that we can reproduce existing results, we perform a study over the subset of this search space on~\textsc{cifar10}. We then proceed to analyze the performance of these models across $\celebahq$ and $\lsun$. In Section~\ref{sec:penalty} we fix everything but the regularization and normalization scheme. In Section~\ref{sec:loss} we fix everything but the loss. Finally, in Section~\ref{sec:architecture} we fix everything but the architecture. This allows us to decouple some of these design choices and provide some insight on what matters most in practice.

\begin{table}[h]
\centering
\begin{tabular}{lllll}
  \toprule
  \textsc{Parameter} & \textsc{Discrete Value} \\\midrule
  Learning rate $\alpha$ & $\{0.0002, 0.0001, 0.001\}$ \\ \midrule
  Reg. strength $\lambda$ & $\{1, 10\}$ \\ \midrule
  $(\beta_1, \beta_2, n_{dis})$ & $\{(0.5, 0.900, 5)$, $(0.5, 0.999, 1),$ \\
                     & \ \ $(0.5, 0.999, 5)$, $(0.9, 0.999, 5)\}$ \\ \bottomrule
\end{tabular}
\caption{Hyperparameter ranges used in this study. The Cartesian product of the fixed values suffices to uncover most of the recent results from the literature. \label{tab:seed_parameters}}
\end{table}
 
\begin{table}[h!]
\centering
\begin{tabular}{lllll}
  \toprule
  \textsc{Parameter} & \textsc{Range}& \textsc{Log} \\\midrule
  Learning rate $\alpha$ & $[10^{-5}, 10^{-2}]$ & Yes \\ \midrule
  $\lambda$ for $L_2$ & $[10^{-4}, 10^1]$ & Yes \\
  $\lambda$ for non-$L_2$ & $[10^{-1}, 10^2]$ & Yes \\ \midrule
  $\beta_1 \times \beta_2$ & $[0, 1] \times [0, 1]$ & No \\ \bottomrule
\end{tabular}
\caption{We use sequential Bayesian optimization~\citep{srinivasKKS10} to explore the hyperparameter settings from the specified ranges. We explore 120 hyperparameter settings in 12 rounds of optimization. \label{tab:vizier_parameter_range}}
\end{table}

\begin{figure*}[t!]
    \centering
    \includegraphics[width=0.9\textwidth]{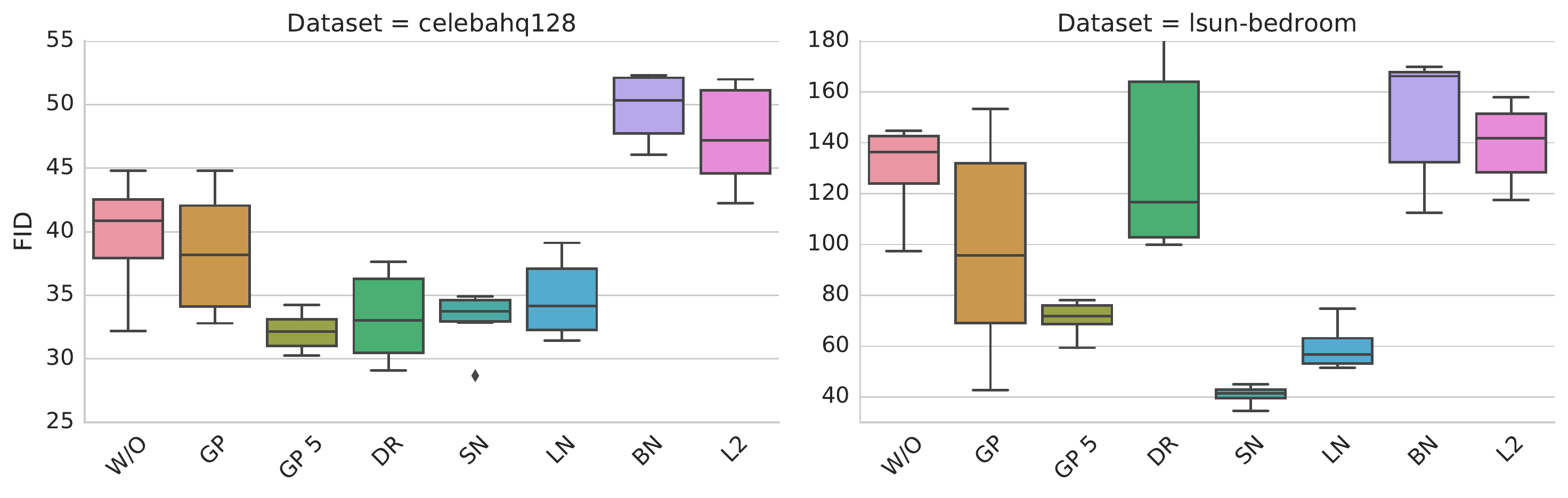}
    \includegraphics[width=0.9\textwidth]{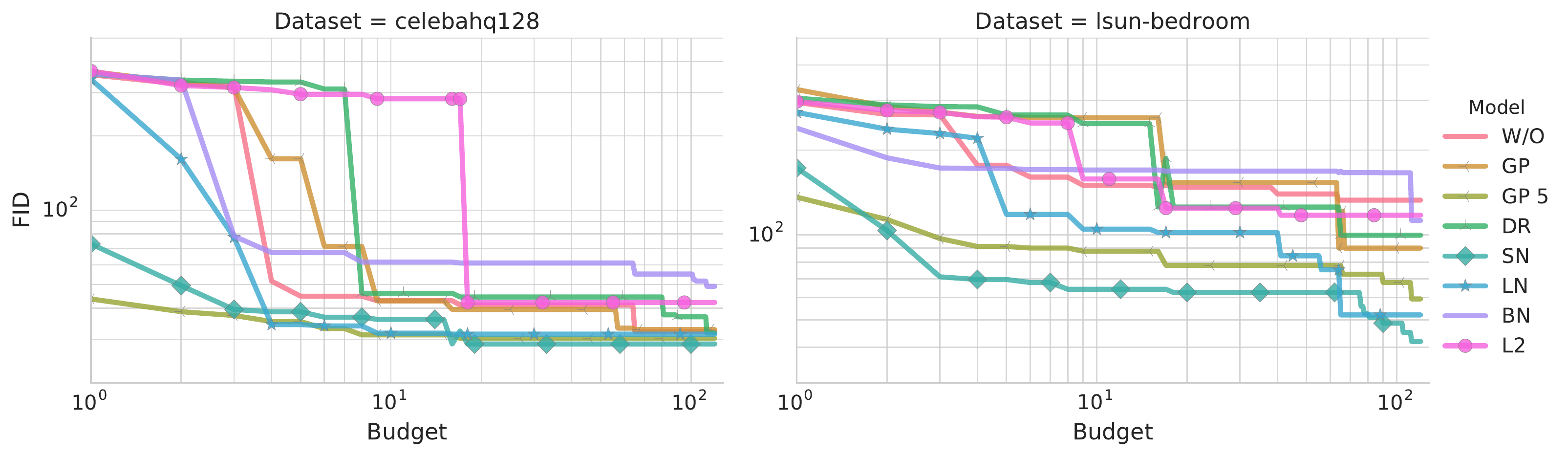}
    \caption{Plots in the first row show the FID distribution for top $5\%$ models (lower is better). We observe that both gradient penalty (GP) and spectral normalization (SN) outperform the non-regularized/normalized baseline (W/O). Unfortunately, none fully address the stability issues. The second row shows the estimate of the minimum FID achievable for a given computational budget. For example, to obtain an FID below 100 using non-saturating loss with gradient penalty, we need to 
    try at least 6 hyperparameter settings. At the same time, we could achieve a better result (lower FID) with spectral normalization and 2 hyperparameter settings. These results suggest that spectral norm is a better practical choice.}
    \label{fig:penalty_top_10}
\end{figure*}

As noted by~\citet{lucic2017gans}, one major issue preventing further progress is the hyperparameter tuning -- currently, the community has converged to a small set of parameter values which work on some datasets, and may completely fail on others. In this study we combine the best hyperparameter settings found in the literature~\citep{miyato2018spectral}, and perform sequential Bayesian optimization~\citep{srinivasKKS10} to possibly uncover better hyperparameter settings. In a nutshell, in sequential Bayesian optimization one starts by evaluating a set of hyperparameter settings (possibly chosen randomly). Then, based on the obtained scores for these hyperparameters the next set of hyperparameter combinations is chosen such to balance the exploration (finding new hyperparameter settings which might perform well) and exploitation (selecting settings close to the best-performing settings). We then consider the top performing models and discuss the impact of the computational budget.

\begin{figure*}[t!]
    \centering
    \includegraphics[width=0.9\textwidth]{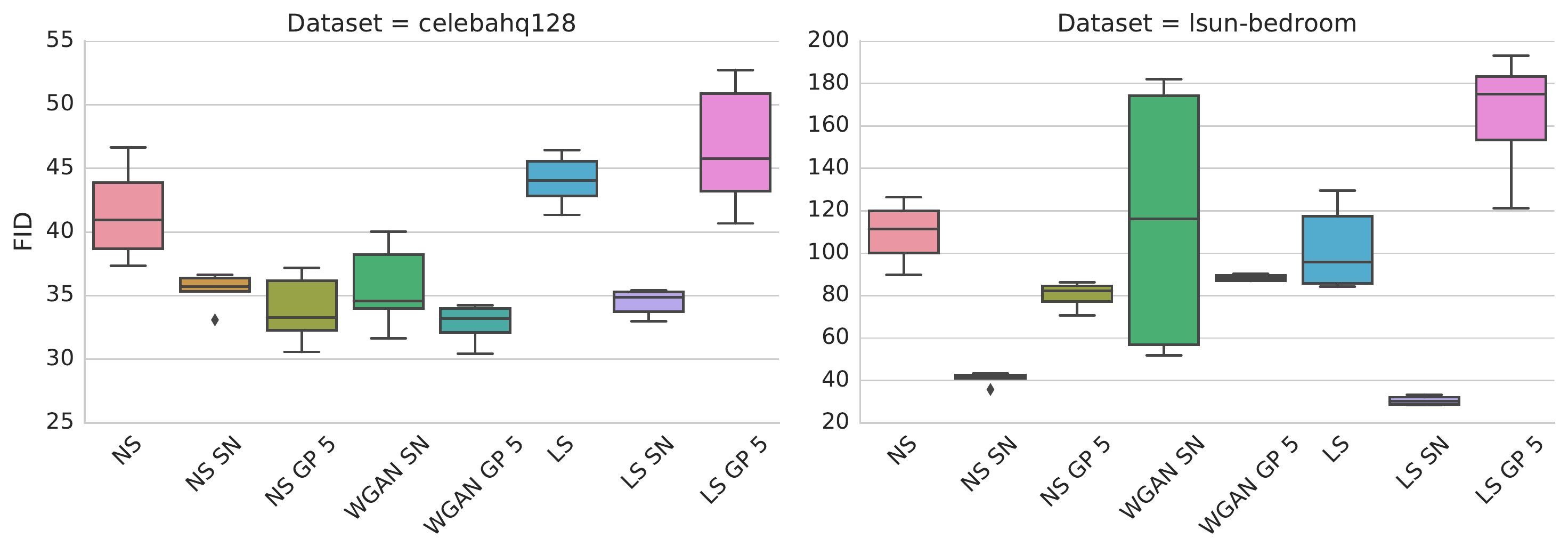}
    \includegraphics[width=0.9\textwidth]{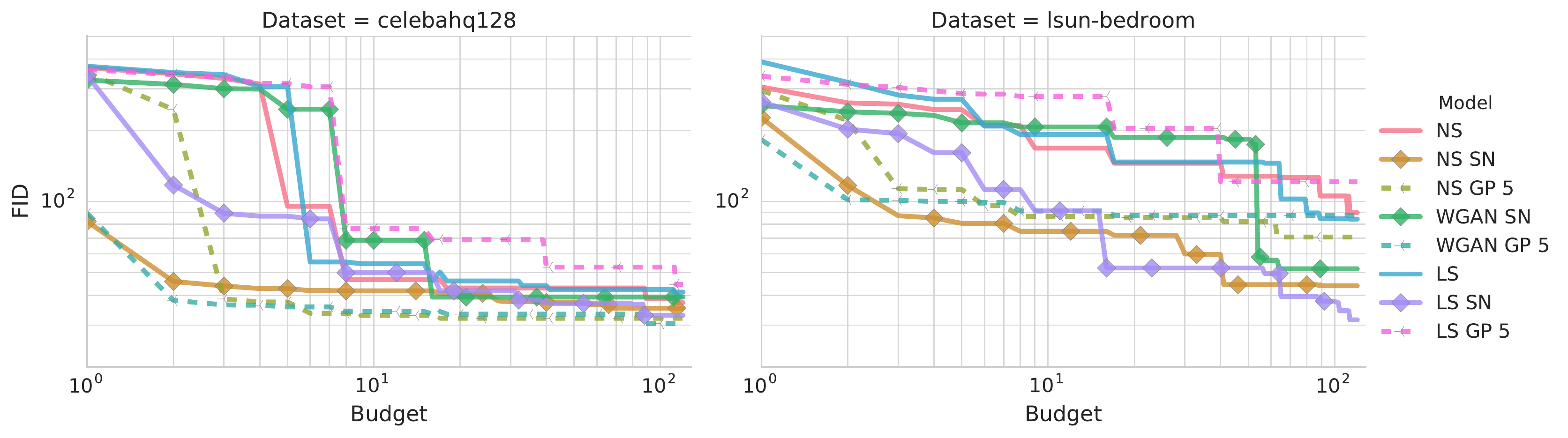}
    \caption{The first row shows the FID distribution for top $5\%$ models. We compare the non-saturating (NS) loss, the Wasserstein loss (WGAN), and the least-squares loss (LS), combined with the most prominent regularization and normalization strategies, namely spectral norm (SN) and gradient penalty (GP). We observe that spectral norm consistently improves the sample quality. In some cases the gradient penalty can help, but there is no clear trend. From the computational budget perspective one can attain lower levels of FID with fewer hyperparameter optimization settings which demonstrates the practical advantages of spectral normalization over competing method.}
    \label{fig:loss_top_10}
\end{figure*}

\looseness=-1We summarize the fixed hyperparameter settings in Table~\ref{tab:seed_parameters} which contains the ``good'' parameters reported in recent publications~\citep{fedus2018many,miyato2018spectral,gulrajani2017improved}. In particular, we consider the Cartesian product of these parameters to obtain 24 hyperparameter settings to reduce the survivorship bias. Finally, to provide a fair comparison, we perform sequential Bayesian optimization~\citep{srinivasKKS10} on the parameter ranges provided in Table~\ref{tab:vizier_parameter_range}. We run $12$ rounds (i.e. we communicate with the oracle 12 times) of sequential optimization, each with a batch of $10$ hyperparameter sets selected based on the FID scores from the results of the previous iterations. As we explore the number of discriminator updates per generator update (1 or 5), this leads to an additional $240$ hyperparameter settings which in some cases outperform the previously known hyperparameter settings. The batch size is set to 64 for all the experiments. We use a fixed the number of discriminator update steps of 100K for $\lsun$ dataset and $\celebahq$ dataset, and 200K for \textsc{cifar10} dataset. We apply the Adam optimizer~\citep{kingma2014adam}.

\section{Experimental Results and Discussion}

Given that there are 4 major components (loss, architecture, regularization, normalization) to analyze for each dataset, it is infeasible to explore the whole landscape. Hence, we opt for a more pragmatic solution -- we keep some dimensions fixed, and vary the others. We highlight two aspects:
\begin{enumerate}[itemsep=0mm,topsep=0mm,parsep=1mm]
    \item We train the models using various hyperparameter settings, both predefined and ones obtained by sequential Bayesian optimization. Then we compute the \emph{FID distribution} of the top $5\%$ of the trained models. The lower the median FID, the better the model. The lower the variance, the more stable the model is from the optimization point of view. 
    \item The tradeoff between the computational budget (for training) and model quality in terms of FID. Intuitively, given a limited computational budget (being able to train only $k$ different models), which model should one choose? Clearly, models which achieve better performance using the same computational budget should be preferred in practice. To compute the minimum attainable FID for a fixed budget $k$ we simulate a practitioner attempting to find a good hyperparameter setting for their model: we spend a part of the budget on the ``good'' hyperparameter settings reported in recent publications, followed by exploring new settings (i.e. using Bayesian optimization). As this is a random process, we repeat it 1000 times and report the \emph{average} of the minimum attainable FID.
\end{enumerate}
    
Due to the fact that the training is sensitive to the initial weights, we train the models 5 times, each time with a different random initialization, and report the median FID. The variance in FID for models obtained by sequential Bayesian optimization is handled implicitly by the applied exploration-exploitation strategy.

\begin{figure*}[t]
    \centering
    \includegraphics[width=0.9\textwidth]{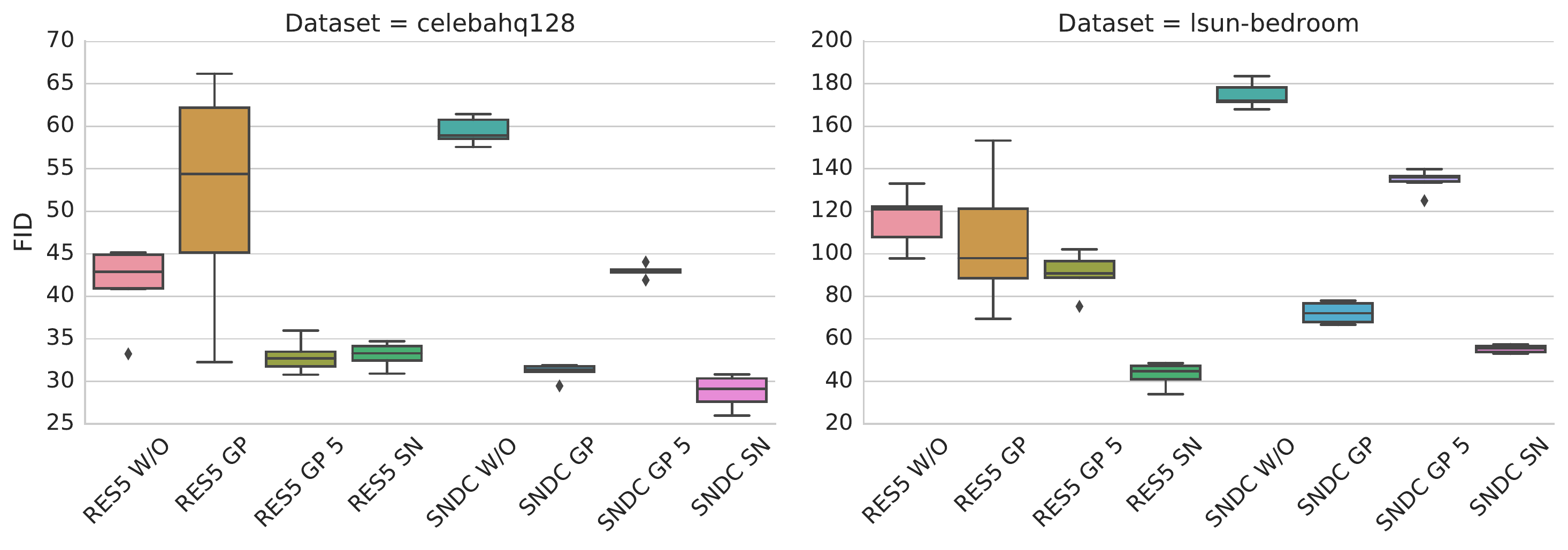}
    \includegraphics[width=0.9\textwidth]{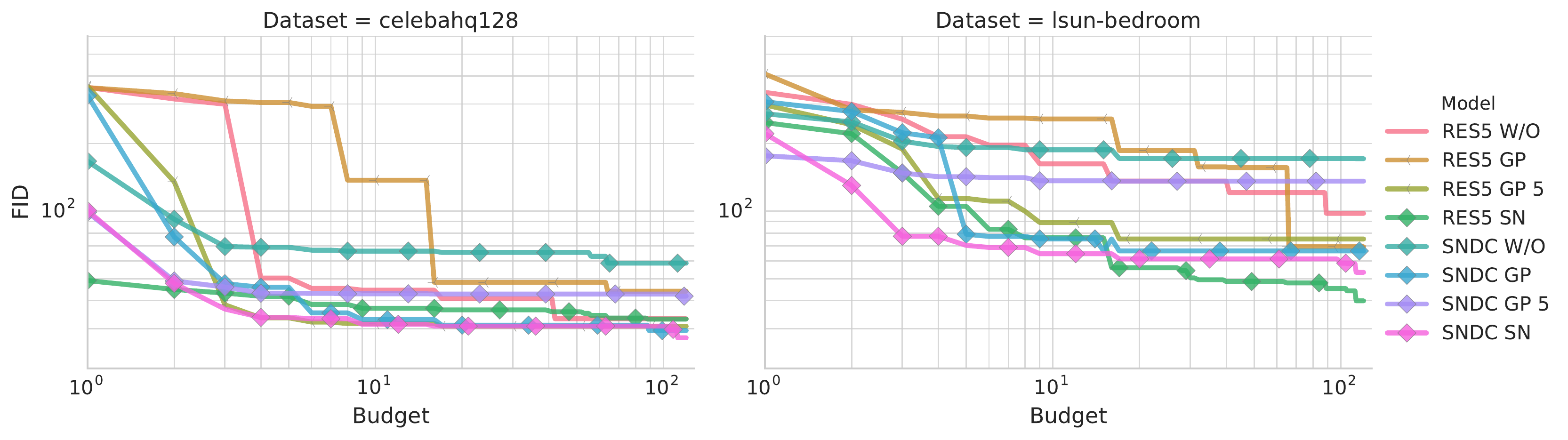}
    \caption{The first row show s the FID distribution for top $5\%$ models. We compare the ResNet-based neural architecture with the SNDCGAN architecture. We use the non-saturating (NS) loss in all experiments, and apply either spectral normalization (SN) or the gradient penalty (GP).  We observe that spectral norm consistently improves the sample quality. In some cases the gradient penalty can help, but the need to tune one additional hyperparameter leads to a lower computational efficiency.}
    \label{fig:architecture_top_5}
\end{figure*}
\subsection{Regularization and Normalization}\label{sec:penalty}
The goal of this study is to compare the relative performance of various regularization and normalization methods presented in the literature, namely:
batch normalization (BN)~\citep{ioffe2015batch}, layer normalization (LN)~\citep{ba2016layer}, spectral normalization (SN), gradient penalty (GP)~\citep{gulrajani2017improved}, Dragan penalty (DR)~\citep{kodali2017dragan}, or $L_2$ regularization. We fix the loss to non-saturating loss~\citep{goodfellow2014generative} and the ResNet19 with generator and discriminator architectures described in Table~\ref{tab:resnet19_discriminator}. We analyze the impact of the loss function in Section~\ref{sec:loss} and of the architecture in Section~\ref{sec:architecture}. We consider both \textsc{CelebA-HQ-128} and \textsc{LSUN-bedroom} with the hyperparameter settings shown in Tables~\ref{tab:seed_parameters} and~\ref{tab:vizier_parameter_range}.

The results are presented in Figure~\ref{fig:penalty_top_10}. We observe that adding batch norm to the discriminator hurts the performance. Secondly, gradient penalty can help, but it doesn't stabilize the training. In fact, it is non-trivial to strike a balance of the loss and regularization strength. Spectral normalization helps improve the model quality and is more computationally efficient than gradient penalty. This is consistent with recent results in~\citet{zhang2018self}. Similarly to the loss study, models using GP penalty may benefit from 5:1 ratio of discriminator to generator updates. Furthermore, in a separate ablation study we observed that running the optimization procedure for an additional 100K steps is likely to increase the performance of the models with GP penalty.

\subsection{Impact of the Loss Function}\label{sec:loss}
Here we investigate whether the above findings also hold when the loss functions are varied. In addition to the non-saturating loss (NS), we also consider the  the least-squares loss (LS)~\citep{mao2016least}, or the Wasserstein loss
(WGAN)~\citep{arjovsky2017wasserstein}. We use the ResNet19 with generator and
discriminator architectures detailed in Table~\ref{tab:resnet19_discriminator}.
We consider the most prominent normalization and regularization approaches: gradient penalty~\citep{gulrajani2017improved}, and spectral normalization~\citep{miyato2018spectral}. Other parameters are detailed in Table~\ref{tab:seed_parameters}. We also performed a study on the recently popularized hinge loss~\citep{lim2017geometric,miyato2018spectral,brock2018large} and present it in the Appendix.

The results are presented in Figure~\ref{fig:loss_top_10}. Spectral normalization improves the model quality on both datasets. Similarly, the gradient penalty can help, but finding a good regularization tradeoff is non-trivial and requires a large computational budget. Models using the GP penalty benefit from 5:1 ratio of discriminator to generator updates~\citep{gulrajani2017improved}. 

\subsection{Impact of the Neural Architectures}\label{sec:architecture}
\looseness=-1An interesting practical question is whether our findings also hold for different neural architectures. To this end, we also perform a study on SNDCGAN from ~\citet{miyato2018spectral}. We consider the non-saturating GAN loss, gradient penalty and spectral normalization. While for smaller architectures regularization is not essential~\citep{lucic2017gans}, the regularization and normalization effects might become more relevant due to deeper architectures and optimization considerations.

The results are presented in Figure~\ref{fig:architecture_top_5}. We observe that both architectures achieve comparable results and benefit from regularization and normalization. Spectral normalization strongly outperforms the baseline for both architectures.

\begin{figure*}[t]
    \centering
    \includegraphics[width=0.9\textwidth]{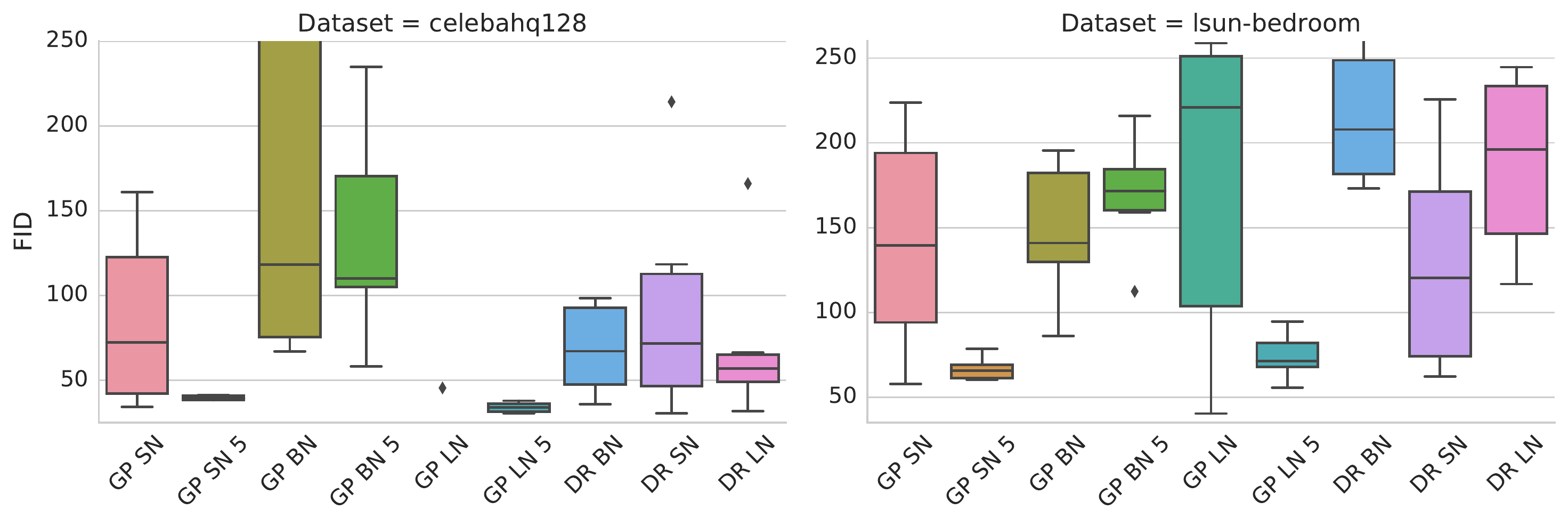}
    \caption{Can one benefit from simultaneous regularization and normalization? The plots show the FID distribution for top $5\%$ models where we compare various combinations of regularization and normalization strategies. Gradient penalty coupled with spectral normalization (SN) or layer normalization (LN) strongly improves the performance over the baseline. This can be partially explained by the fact that SN doesn't ensure that the discriminator is 1-Lipschitz due to the way convolutional layers are normalized.}
    \label{fig:penalty_combo_top_10}
\end{figure*}

\looseness=-1\textbf{Simultaneous Regularization and Normalization}\quad
A common observation is that the Lipschitz constant of the discriminator is critical for the performance, one may expect simultaneous regularization and normalization could improve model quality. To quantify this effect, we fix the loss to non-saturating loss~\citep{goodfellow2014generative}, use the Resnet19 architecture (as above), and combine several normalization and regularization schemes, with hyperparameter settings shown in Table~\ref{tab:seed_parameters} coupled with 24 randomly selected parameters. The results are presented in Figure~\ref{fig:penalty_combo_top_10}. We observe that one may benefit from additional regularization and normalization. However, a lot of computational effort has to be invested for somewhat marginal gains in FID. Nevertheless, given enough computational budget we advocate simultaneous regularization and normalization -- spectral normalization and layer normalization seem to perform well in practice.

\section{Challenges of a Large-Scale Study}\label{sec:pitfalls}

In this section we focus on several pitfalls we encountered while trying to reproduce existing results and provide a fair and accurate comparison.

\textbf{Metrics}\quad There already seems to be a divergence in how the FID score is computed: (1) Some authors report the score on training data, yielding a FID between $50$K training and  $50$K generated samples~\citep{unterthiner2018coulomb}. Some opt to report the FID based on $10$K test samples and $5$K generated samples and use a custom implementation~\citep{miyato2018spectral}. Finally,~\citet{lucic2017gans} report the score with respect to the test data, in particular FID between $10$K test samples, and $10$K generated samples. The subtle differences will result in a mismatch between the reported FIDs, in some cases of more than $10\%$. We argue that FID should be computed with respect to the test dataset. Furthermore, whenever possible, one should use the same number of instances as previously reported results. Towards this end we use $10$K test samples and $10$K generated samples on \textsc{cifar10} and $\lsun$, and $3$K vs $3$K on $\celebahq$ as in in~\citet{lucic2017gans}. 

\textbf{Details of Neural Architectures}\quad Even in popular architectures, like ResNet, there is still a number of design decisions one needs to make, that are often omitted from the reported results. Those include the exact design of the ResNet block (order of layers, when is ReLu applied, when to upsample and downsample, how many filters to use). Some of these differences might lead to potentially unfair comparison. As a result, we suggest to use the architectures presented within this work as a solid baseline. An ablation study on various ResNet modifications is available in the Appendix.

\textbf{Datasets}\quad A common issue is related to dataset processing -- does $\lsun$ always correspond to the same dataset? In most cases the precise algorithm for upscaling or cropping is not clear which introduces inconsistencies between results on the ``same'' dataset.

\textbf{Implementation Details and Non-Determinism}\quad One major issue is the mismatch between the algorithm presented in a paper and the code provided online. We are aware that there is an embarrassingly large gap between a good implementation and a bad implementation of a given model. Hence, when no code is available, one is forced to guess which modifications were done. Another particularly tricky issue is removing randomness from the training process. After one fixes the data ordering and the initial weights, obtaining the same score by training the same model twice is non-trivial due to randomness present in certain GPU operations~\citep{chetlur2014cudnn}. Disabling the optimizations causing the non-determinism often results in an order of magnitude running time penalty.

While each of these issues taken in isolation seems minor, they compound to create a mist which introduces friction in practical applications and the research process~\citep{sculley2018winner}.

\section{Related Work}\label{sec:relwork}
A recent large-scale study on GANs and Variational Autoencoders was presented in~\citet{lucic2017gans}. The authors consider several loss functions and regularizers, and study the effect of the loss function on the FID score, with low-to-medium complexity datasets (\textsc{MNIST}, \textsc{cifar10}, \textsc{CelebA}), and a single neural network architecture. In this limited setting, the authors found that there is no statistically significant difference between recently introduced models and the original non-saturating GAN. A study of the effects of gradient-norm regularization in GANs was recently presented in~\citet{fedus2018many}. The authors posit that the gradient penalty can also be applied to the non-saturating GAN, and that, to a limited extent, it reduces the sensitivity to hyperparameter selection. In a recent work on spectral normalization, the authors perform a small study of the competing regularization and normalization approaches~\citep{miyato2018spectral}. We are happy to report that we could reproduce these results and we present them in the Appendix.

Inspired by these works and building on the available open-source code from~\citet{lucic2017gans}, we take one additional step in all dimensions considered therein: more complex neural architectures, more complex datasets, and more involved regularization and normalization schemes.

\section{Conclusions and Future Work}
In this work we study the impact of regularization and normalization schemes on GAN training. We consider the state-of-the-art approaches and vary the loss functions and neural architectures. We study the impact of these design choices on the quality of generated samples which we assess by recently introduced quantitative metrics. 

Our fair and thorough empirical evaluation suggests that when the computational budget is limited one should consider non-saturating GAN loss and spectral normalization as default choices when applying GANs to a new dataset. Given additional computational budget, we suggest adding the gradient penalty from~\citet{gulrajani2017improved} and training the model until convergence. Furthermore, we observe that both classes of popular neural architectures can perform well across the considered datasets. A separate ablation study uncovered that most of the variations applied in the ResNet style architectures lead to marginal improvements in the sample quality.

As a result of this large-scale study we identify the common pitfalls standing in the way of accurate and fair comparison and propose concrete actions to demystify the future results -- issues with metrics, dataset preprocessing, non-determinism, and missing implementation details are particularly striking. We hope that this work, together with the open-sourced reference implementations and trained models, will serve as a solid baseline for future GAN research.

Future work should carefully evaluate models which necessitate large-scale training such as BigGAN~\citep{brock2018large}, models with custom architectures~\citep{chen2018selfmodulation,karras2018style,zhang2018self}, recently proposed regularization techniques~\citep{roth2017stabilizing, mescheder2018training}, and other proposals for stabilizing the training~\citep{chen2018self}. In addition, given the popularity of conditional GANs, one should explore whether these insights transfer to the conditional settings. Finally, given the drawbacks of FID and IS, additional quantitative evaluation using recently proposed metrics could bring novel insights~\citep{sajjadi2018assessing,kynkaanniemi2019improved}.

\section*{Acknowledgments}
We are grateful to Michael Tschannen for detailed comments on this manuscript.

\bibliographystyle{icml2019}
\bibliography{paper.bib}

\newpage
\onecolumn

\appendix

\section{FID and Inception Scores on CIFAR10}\label{sec:repro}

\begin{figure}[ht]
  \centering
  \includegraphics[width=.75\textwidth]{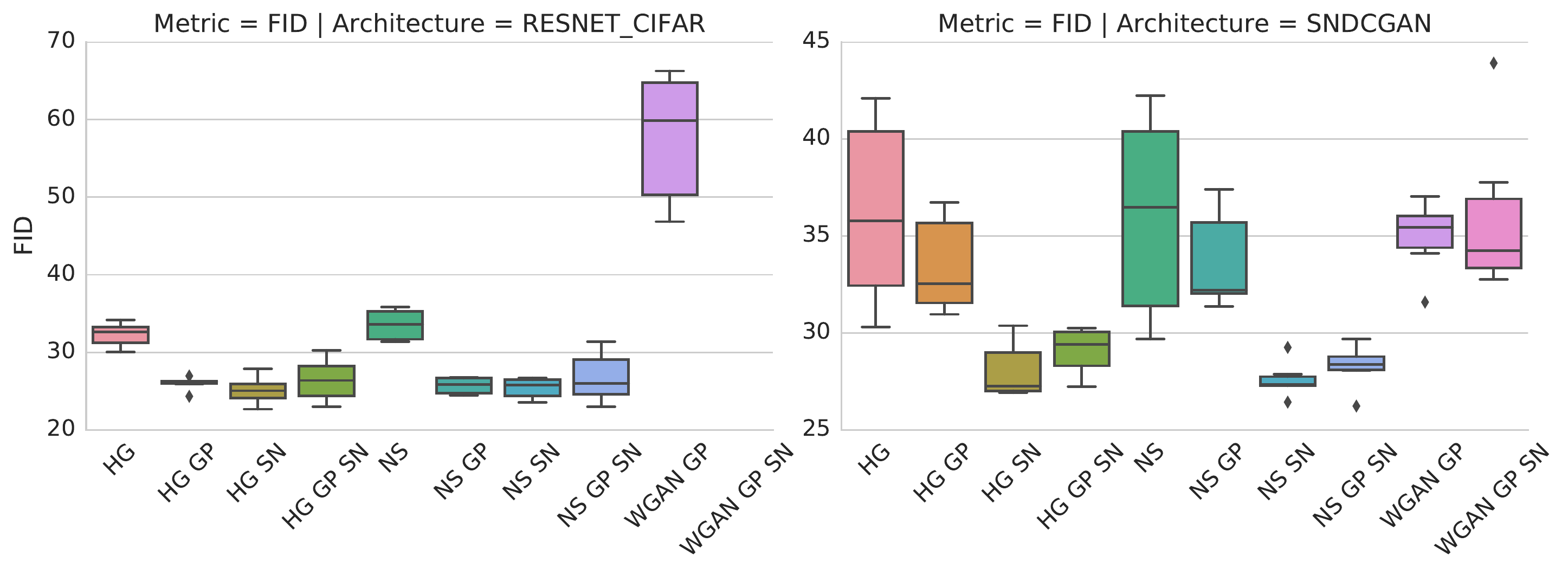}
  \caption{An empirical study with SNDCGAN and ResNet CIFAR architectures on \textsc{cifar10}. We recover the results reported in~\citet{miyato2018spectral}.}
  \label{fig:reproduction_fid}
\end{figure}

\begin{figure}[ht]
    \centering
    \includegraphics[width=.8\textwidth]{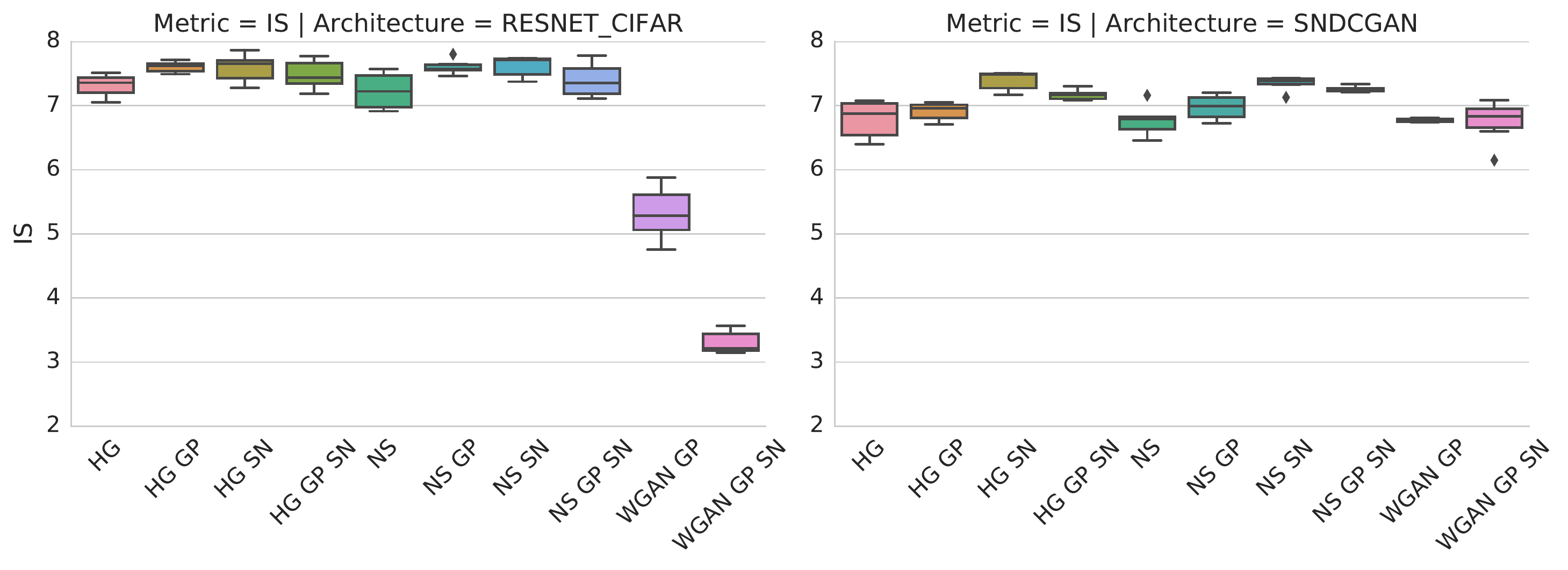}
    \caption{Inception Score for each model within our study which corresponds results reported in~\citet{miyato2018spectral}.}
    \label{fig:reproduction_is}
\end{figure}

We present an empirical study with SNDCGAN and ResNet CIFAR architectures on
\textsc{cifar10} in figure~\ref{fig:reproduction_fid} and
figure~\ref{fig:reproduction_is}. In addition to the non-saturating loss (NS) and the Wasserstein loss (WGAN) presented in Section~\ref{sec:loss}, we evaluate hinge loss (HG) on~\textsc{cifar10}. We observe that its performance is similar to the non-saturating loss.

\section{Empirical Comparison of FID and KID}\label{sec:fid_kid_comparison}

The KID metric introduced by~\citet{binkowski2018demystifying} is an
alternative to FID. We use models from our Regularization and Normalization study (see
Section~\ref{sec:penalty}) to compare both metrics. Here, by model we denote
everything that needs to be specified for the training -- including all
hyper-parameters, like learning rate, $\lambda$, Adam's $\beta$, etc.
The Spearman rank-order correlation coefficient between KID and FID scores is approximately $0.994$ for
$\lsun$ and $0.995$ for $\celebahq$ datasets.

To evaluate a practical setting of selecting several best models, we compare the
intersection between the set of ``best $K$ models by FID" and the set of
``best $K$ models by KID" for $K \in {5, 10, 20, 50, 100}$. The results are
summarized in Table~\ref{tab:fid_vs_kid}.

This experiment suggests that FID and KID metrics are very strongly correlated, and for the
practical applications one can choose either of them. Also, the
conclusions from our studies based on FID should transfer to studies based on KID.

\begin{table}[h!]
\centering
\caption{Intersection between set of top $K$ experiments selected by FID and KID
  metrics.}

\begin{tabular}{lcc}
  \toprule
   & $\lsun$ & $\celebahq$ \\
  \toprule
  \textsc{K = 5} & $4 / 5$ & $2 / 5$ \\
  \textsc{K = 10} & $9 / 10$  & $8 / 10$ \\
  \textsc{K = 20} & $18 / 20$ & $15 / 20$ \\
  \textsc{K = 50} & $49 / 50$ & $46 / 50$ \\
  \textsc{K = 100} & $95 / 100$ & $98 / 100$ \\
  \bottomrule
\end{tabular}
\label{tab:fid_vs_kid}
\end{table}

\newpage
\section{Architecture Details}
\label{sec:architecture_details}

\subsection{SNDCGAN}
We used the same architecture as \citet{miyato2018spectral}, with the parameters copied from the GitHub page\footnote{\url{github.com/pfnet-research/chainer-gan-lib}}. In Table~\ref{tab:sndcgan_discriminator} and Table~\ref{tab:sndcgan_generator},
we describe the operations in layer column with order. Kernel size is described in format $[filter{\_}h, filter{\_}w, stride]$,
input shape is $h \times w$ and output shape is $h \times w \times channels$.
The slopes of all lReLU functions are set to 0.1. The input shape $h \times w$ is
$128\times 128$ for \textsc{celeba-hq-128} and $\lsun$,
$32\times 32$ for \textsc{cifar10}.

\begin{table}[!htb]
    \caption{SNDCGAN architecture.}
    \begin{subtable}[t]{.5\linewidth}
      \centering
      \small
        \caption{SNDCGAN discriminator}
        \begin{tabular}{lll}
          \toprule
          \textsc{Layer} & \textsc{Kernel}& \textsc{Output} \\\toprule
          Conv, lReLU & $[3, 3, 1]$ & $h \times w \times 64$  \\ \midrule
          Conv, lReLU & $[4, 4, 2]$ & $h/2 \times w/2 \times 128$ \\ \midrule
          Conv, lReLU & $[3, 3, 1]$ & $h/2 \times w/2 \times 128$ \\ \midrule
          Conv, lReLU & $[4, 4, 2]$ & $h/4 \times w/4 \times 256$ \\ \midrule
          Conv, lReLU & $[3, 3, 1]$ & $h/4 \times w/4 \times 256$ \\ \midrule
          Conv, lReLU & $[4, 4, 2]$ & $h/8 \times w/8 \times 512$ \\ \midrule
          Conv, lReLU & $[3, 3, 1]$ & $h/8 \times w/8 \times 512$ \\ \midrule
          Linear & - & 1 \\ \bottomrule
        \end{tabular}
        \label{tab:sndcgan_discriminator}
    \end{subtable}%
    \begin{subtable}[t]{.5\linewidth}
      \centering
      \small
        \caption{SNDCGAN generator}
        \begin{tabular}{lll}
          \toprule
          \textsc{Layer} & \textsc{Kernel}& \textsc{Output} \\\toprule
          $z$ & - & $128$  \\ \midrule
          Linear, BN, ReLU & - & $h/8 \times w/8 \times 512$  \\ \midrule
          Deconv, BN, ReLU & $[4, 4, 2]$ & $h/4 \times w/4 \times 256$ \\ \midrule
          Deconv, BN, ReLU & $[4, 4, 2]$ & $h/2 \times w/2 \times 128$ \\ \midrule
          Deconv, BN, ReLU & $[4, 4, 2]$ & $h \times w \times 64$ \\ \midrule
          Deconv, Tanh & $[3, 3, 1]$ & $h \times w \times 3$ \\ \bottomrule
        \end{tabular}
        \label{tab:sndcgan_generator}
    \end{subtable}
\label{tab:sndcgan_architecture}
\end{table}

\subsection{ResNet Architecture}
The ResNet19 architecture is described in Table~\ref{tab:resnet19_architecture}.
The RS column stands for the resample of the residual block, with downscale(D)/upscale(U)/none(-) setting. MP stands for mean pooling and BN for batch normalization. ResBlock is defined in Table~\ref{tab:resnet_block}. The addition layer merges two paths by adding them. The first path is a shortcut layer with exactly one convolution operation, while the second path consists of two convolution operations. The downscale layer and upscale layer are marked in
Table~\ref{tab:resnet_block}. We used average pool with kernel $[2, 2, 2]$ for downscale, after the convolution operation. We used unpool from \url{github.com/tensorflow/tensorflow/issues/2169} for upscale, before the convolution operation. $h$ and $w$ are the input shape to the ResNet block, output
shape depends on the RS parameter. $c_{i}$ and $c_{o}$ are the input channels
and output channels for a ResNet block. Table~\ref{tab:resnet_cifar_architecture} described the ResNet CIFAR architecture we used in Figure~\ref{fig:reproduction_fid} for reproducing the existing results. Note
that RS is set to none for third ResBlock and fourth ResBlock in discriminator.
In this case, we used the same ResNet block defined in
Table~\ref{tab:resnet_block} without resampling.

\newpage  
\begin{table}[!htb]
    \caption{ResNet 19 architecture corresponding to ``resnet\_small" in \url{github.com/pfnet-research/sngan_projection}.}
    \begin{subtable}[t]{.5\linewidth}
      \centering
      \small
        \caption{ResNet19 discriminator}
        \begin{tabular}{llll}
          \toprule
          \textsc{Layer} & \textsc{Kernel}& \textsc{RS} & \textsc{Output} \\\toprule
          ResBlock & $[3,3,1]$ & D & $64 \times 64 \times 64$ \\ \midrule
          ResBlock & $[3,3,1]$ & D & $32 \times 32 \times 128$ \\ \midrule
          ResBlock & $[3,3,1]$ & D & $16 \times 16 \times 256$ \\ \midrule
          ResBlock & $[3,3,1]$ & D & $8 \times 8 \times 256$ \\ \midrule
          ResBlock & $[3,3,1]$ & D & $4 \times 4 \times 512$ \\ \midrule
          ResBlock & $[3,3,1]$ & D & $2 \times 2 \times 512$ \\ \midrule
          ReLU, MP & - & - & $512$ \\ \midrule
          Linear & - & - & 1 \\ \bottomrule
        \end{tabular}
        \label{tab:resnet19_discriminator}
    \end{subtable}%
    \begin{subtable}[t]{.5\linewidth}
      \centering
      \small
        \caption{ResNet19 generator}
        \begin{tabular}{llll}
          \toprule
          \textsc{Layer} & \textsc{Kernel}& \textsc{RS} & \textsc{Output} \\\toprule
          $z$ & - & - & $128$ \\ \midrule
          Linear & - & - & $4 \times 4 \times 512$  \\ \midrule
          ResBlock & $[3,3,1]$ & U & $8 \times 8 \times 512$ \\ \midrule
          ResBlock & $[3,3,1]$ & U & $16 \times 16 \times 256$ \\ \midrule
          ResBlock & $[3,3,1]$ & U & $32 \times 32 \times 256$ \\ \midrule
          ResBlock & $[3,3,1]$ & U & $64 \times 64 \times 128$ \\ \midrule
          ResBlock & $[3,3,1]$ & U & $128 \times 128 \times 64$ \\ \midrule
          BN, ReLU & - & - & $128 \times 128 \times 64$ \\ \midrule
          Conv & $[3,3,1]$ & - & $128 \times 128 \times 3$  \\ \midrule
          Sigmoid & - & - & $128 \times 128 \times 3$  \\ \bottomrule
        \end{tabular}
        \label{tab:resnet19_generator}
    \end{subtable}
\label{tab:resnet19_architecture}
\end{table}

\begin{table}[!h]
    \caption{ResNet block definition.}
    \begin{subtable}[t]{.5\linewidth}
      \centering
      \small
        \caption{ResBlock discriminator}
        \begin{tabular}{llll}
          \toprule
          \textsc{Layer} & \textsc{Kernel}& \textsc{RS} & \textsc{Output} \\\toprule
          Shortcut & $[3,3,1]$ & D & $h/2 \times w/2 \times c_{o}$ \\ \midrule
          BN, ReLU & - & - & $h \times w \times c_{i}$ \\
          Conv & $[3,3,1]$ & - & $h \times w \times c_{o}$ \\
          BN, ReLU & - & - & $h \times w \times c_{o}$ \\
          Conv & $[3,3,1]$ & D & $h/2 \times w/2 \times c_{o}$ \\ \midrule
          Addition & - & - & $h/2 \times w/2 \times c_{o}$ \\ \bottomrule
        \end{tabular}
        \label{tab:resblock_discriminator}
    \end{subtable}%
    \begin{subtable}[t]{.5\linewidth}
      \centering
      \small
        \caption{ResBlock generator}
        \begin{tabular}{llll}
          \toprule
          \textsc{Layer} & \textsc{Kernel}& \textsc{RS} & \textsc{Output} \\\toprule
          Shortcut & $[3,3,1]$ & U & $2h \times 2w \times c_{o}$ \\ \midrule
          BN, ReLU & - & - & $h \times w \times c_{i}$ \\
          Conv & $[3,3,1]$ & U & $2h \times 2w \times c_{o}$ \\
          BN, ReLU & - & - & $2h \times 2w \times c_{o}$ \\
          Conv & $[3,3,1]$ & - & $2h \times 2w \times c_{o}$ \\ \midrule
          Addition & - & - & $2h \times 2w \times c_{o}$ \\ \bottomrule
        \end{tabular}
        \label{tab:resblock_generator}
    \end{subtable}
\label{tab:resnet_block}
\end{table}

\begin{table}[!htb]
    \caption{ResNet CIFAR architecture.}
    \begin{subtable}[t]{.5\linewidth}
      \centering
      \small
        \caption{ResNet CIFAR discriminator}
        \begin{tabular}{llll}
          \toprule
          \textsc{Layer} & \textsc{Kernel}& \textsc{RS} & \textsc{Output} \\\toprule
          ResBlock & $[3,3,1]$ & D & $16 \times 16 \times 128$ \\ \midrule
          ResBlock & $[3,3,1]$ & D & $8 \times 8 \times 128$ \\ \midrule
          ResBlock & $[3,3,1]$ & - & $8 \times 8 \times 128$ \\ \midrule
          ResBlock & $[3,3,1]$ & - & $8 \times 8 \times 128$ \\ \midrule
          ReLU, MP & - & - & $128$ \\ \midrule
          Linear & - & - & 1 \\ \bottomrule
        \end{tabular}
        \label{tab:resnet_cifar_discriminator}
    \end{subtable}%
    \begin{subtable}[t]{.5\linewidth}
      \centering
      \small
        \caption{ResNet CIFAR generator}
        \begin{tabular}{llll}
          \toprule
          \textsc{Layer} & \textsc{Kernel}& \textsc{RS} & \textsc{Output} \\\toprule
          $z$ & - & - & $128$ \\ \midrule
          Linear & - & - & $4 \times 4 \times 256$  \\ \midrule
          ResBlock & $[3,3,1]$ & U & $8 \times 8 \times 256$ \\ \midrule
          ResBlock & $[3,3,1]$ & U & $16 \times 16 \times 256$ \\ \midrule
          ResBlock & $[3,3,1]$ & U & $32 \times 32 \times 256$ \\ \midrule
          BN, ReLU & - & - & $32 \times 32 \times 256$ \\ \midrule
          Conv & $[3,3,1]$ & - & $32 \times 32 \times 3$  \\ \midrule
          Sigmoid & - & - & $32 \times 32 \times 3$  \\ \bottomrule
        \end{tabular}
        \label{tab:resnet_cifar_generator}
    \end{subtable}
\label{tab:resnet_cifar_architecture}
\end{table}

\newpage
\section{ResNet Architecture Ablation Study}
\label{sec:resnet_ablation_study}
\begin{figure}[h]
  \centering
	\begin{subfigure}{0.4\textwidth}
		\includegraphics[width=\textwidth]{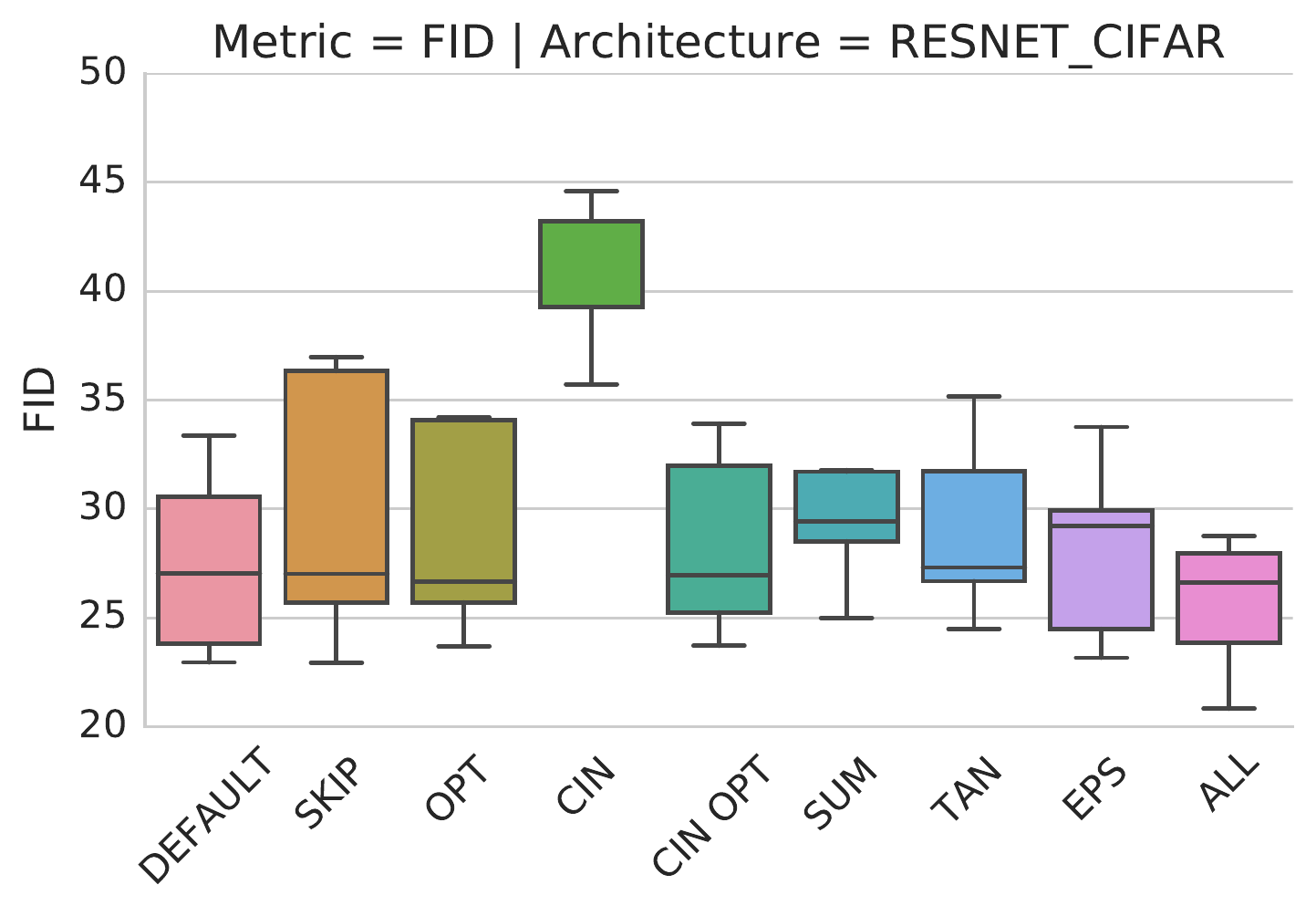}
	\end{subfigure}
	\begin{subfigure}{0.4\textwidth}
		\includegraphics[width=\textwidth]{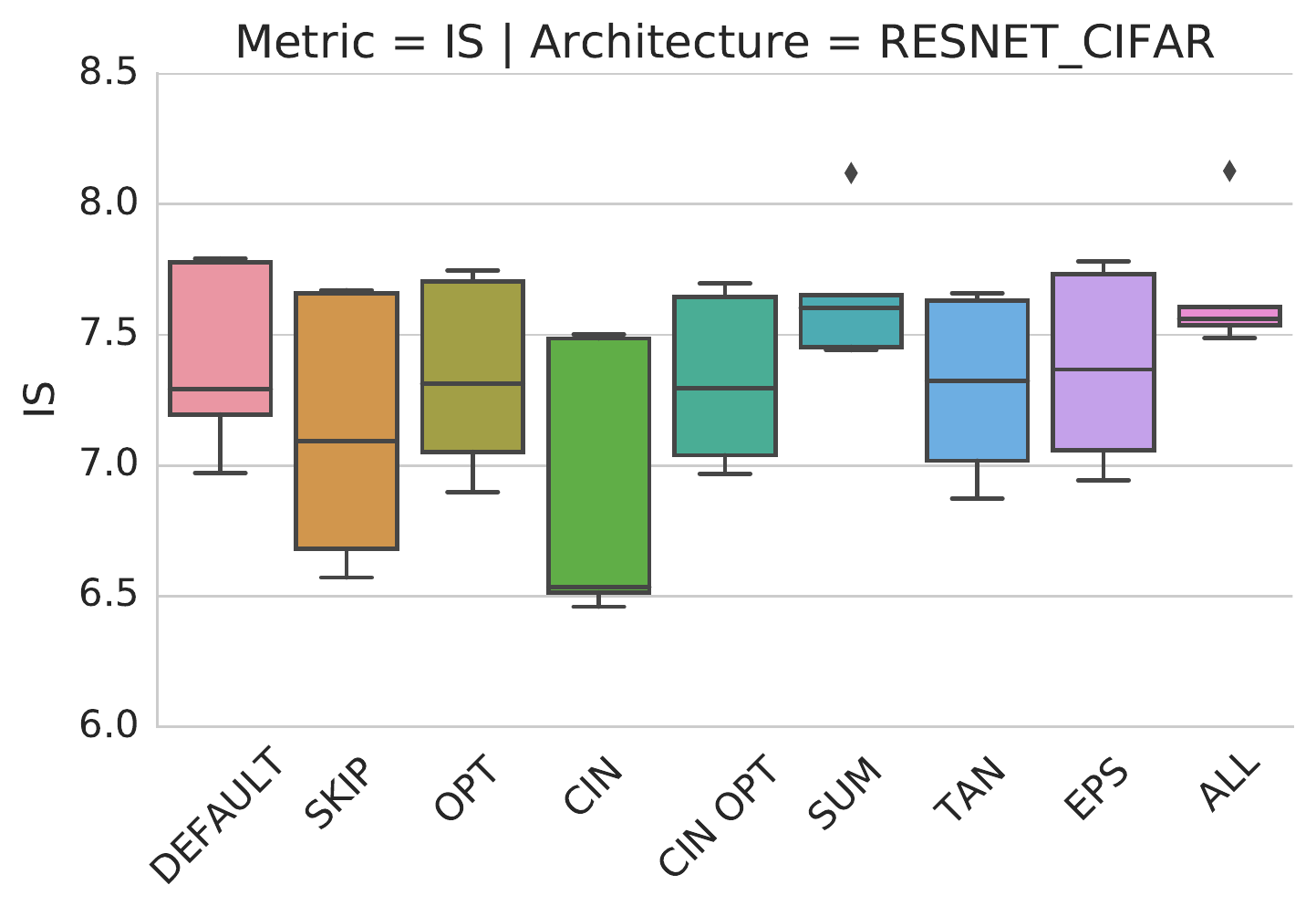}
	\end{subfigure}
  \caption{Ablation study of ResNet architecture differences. The experiment codes are described in Section~\ref{sec:resnet_ablation_study}.}
  \label{fig:resnet_ablation}
\end{figure}

We have noticed six minor differences in the Resnet architecture compared to the implementation from \url{github.com/pfnet-research/chainer-gan-lib/blob/master/common/net.py}~\citep{miyato2018spectral}. We performed an ablation study to verify the impact of these differences. Figure~\ref{fig:resnet_ablation} shows the impact of the ablation study, with details described in the following.

\begin{itemize}
  \item DEFAULT: ResNet CIFAR architecture with spectral normalization and non-saturating GAN loss.
  \item SKIP: Use input as output for the shortcut connection in the discriminator ResBlock. By default it was a convolutional layer with 3x3 kernel.
  \item CIN: Use $c_{i}$ for the discriminator ResBlock hidden layer output channels. By default it was $c_{o}$ in our setup, while~\citet{miyato2018spectral} used $c_{o}$ for first ResBlock and $c_{i}$ for the rest.
  \item OPT: Use an optimized setup for the first discriminator ResBlock, which includes: (1) no ReLU, (2) a convolutional layer for the shortcut connections, (3) use $c_{o}$ instead of $c_{i}$ in ResBlock.
  \item CIN OPT: Use CIN and OPT together. It means the first ResBlock is optimized while the remaining ResBlocks use $c_{i}$ for the hidden output channels.
  \item SUM: Use reduce sum to pool the discriminator output. By default it was reduce mean.
  \item TAN: Use tanh for the generator output, as well as range [-1, 1] for the discriminator input. By default it was sigmoid and discriminator input range $[0, 1]$.
  \item EPS: Use a bigger epsilon $2\mathrm{e}-5$ for generator batch normalization. By default it was $1\mathrm{e}-5$ in TensorFlow.
  \item ALL: Apply all the above differences together.
\end{itemize}

In the ablation study, the CIN experiment obtained the worst FID score. Combining with OPT, the CIN results were improved to the same level as the others which is reasonable because the first block has three input channels, which becomes a bottleneck for the optimization. Hence, using OPT and CIN together performs as well as the others. Overall, the impact of these differences are minor according to the study on \textsc{cifar10}.

\section{Recommended Hyperparameter Settings}
\label{sec:recommended_parameters}
To make the future GAN training simpler, we propose a set of best parameters for
three setups: (1) Best parameters without any regularizer. (2) Best parameters
with only one regularizer. (3) Best parameters with at most two regularizers.
Table~\ref{tab:sndcgan_parameters}, Table~\ref{tab:resnet19_parameters} and
Table~\ref{tab:resnet_cifar_parameters} summarize the top 2 parameters for
SNDCGAN architecture, ResNet19 architecture and ResNet CIFAR architecture,
respectively. Models are ranked according to the median FID score of
five different random seeds with fixed hyper-parameters in
Table~\ref{tab:seed_parameters}.
Note that ranking models according to the best
FID score of different seeds will achieve better but unstable result.
Sequential Bayesian optimization hyper-parameters are not included in this table.
For ResNet19 architecture with at most two regularizers, we have run it only
once due to computational overhead. To show the model stability, we listed
the best FID score out of five seeds from the same parameters in column best.
Spectral normalization is clearly outperforms the other normalizers on SNDCGAN
and ResNet CIFAR architectures, while on ResNet19 both layer normalization and
spectral normalization work well.

\begin{figure}[h]
  \centering
	\begin{subfigure}{0.32\textwidth}
		\includegraphics[width=\textwidth]{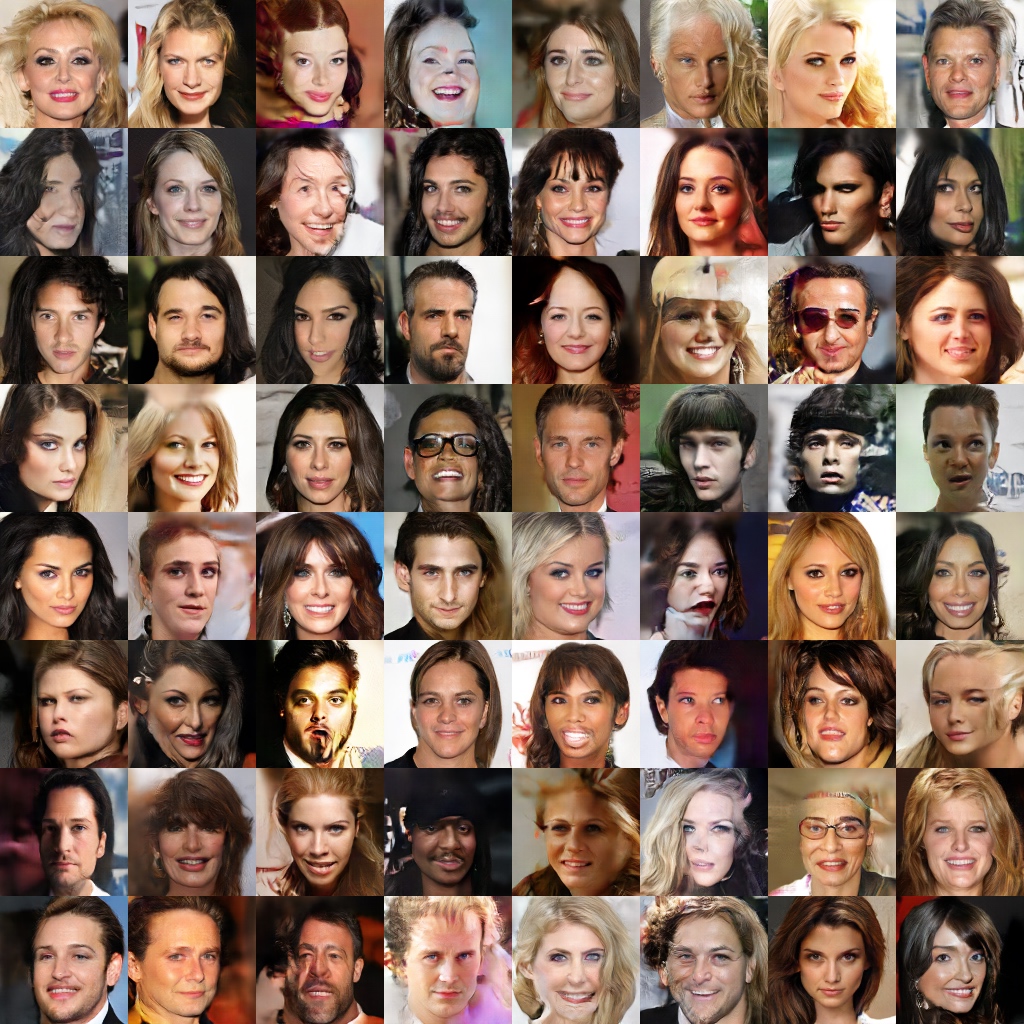}
		\caption{FID = 24.7} 
	\end{subfigure}
	\begin{subfigure}{0.32\textwidth}
		\includegraphics[width=\textwidth]{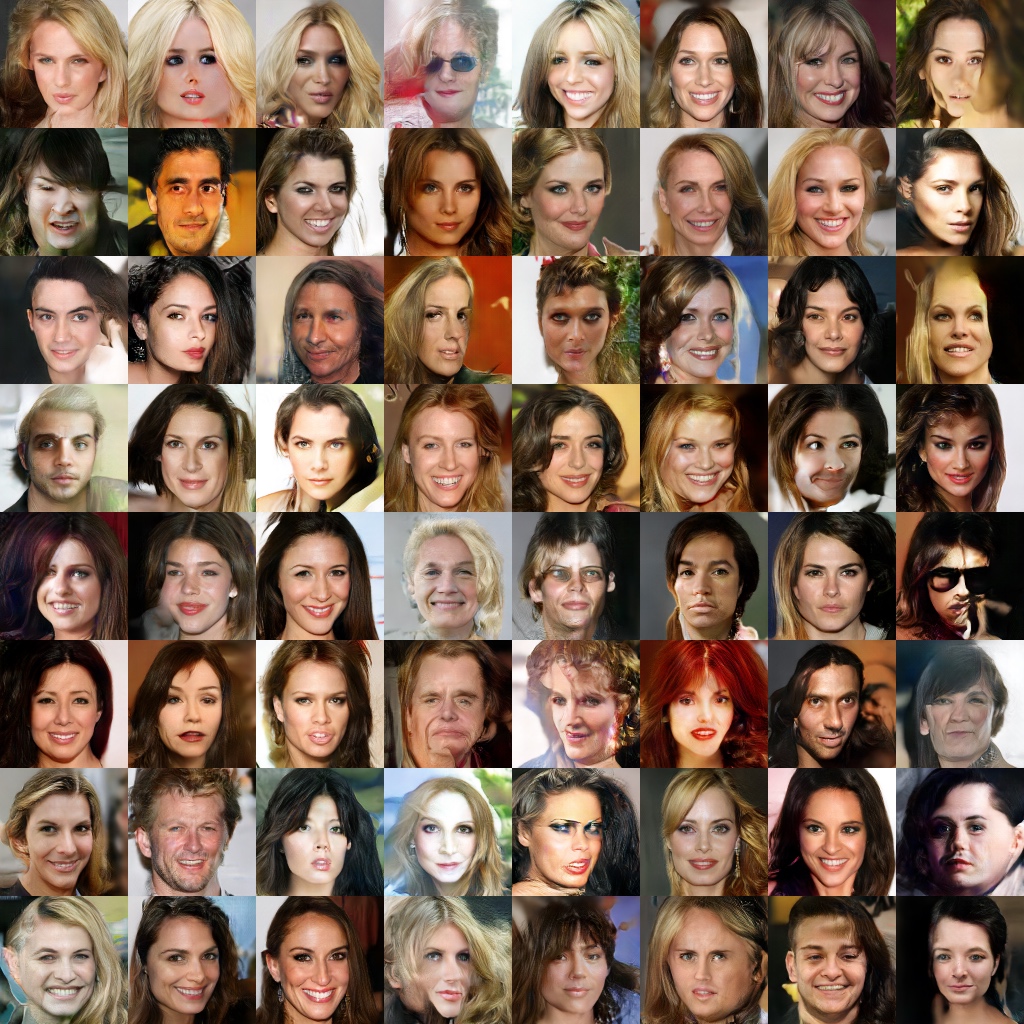}
		\caption{FID = 34.6} 
	\end{subfigure}
	\begin{subfigure}{0.32\textwidth}
		\includegraphics[width=\textwidth]{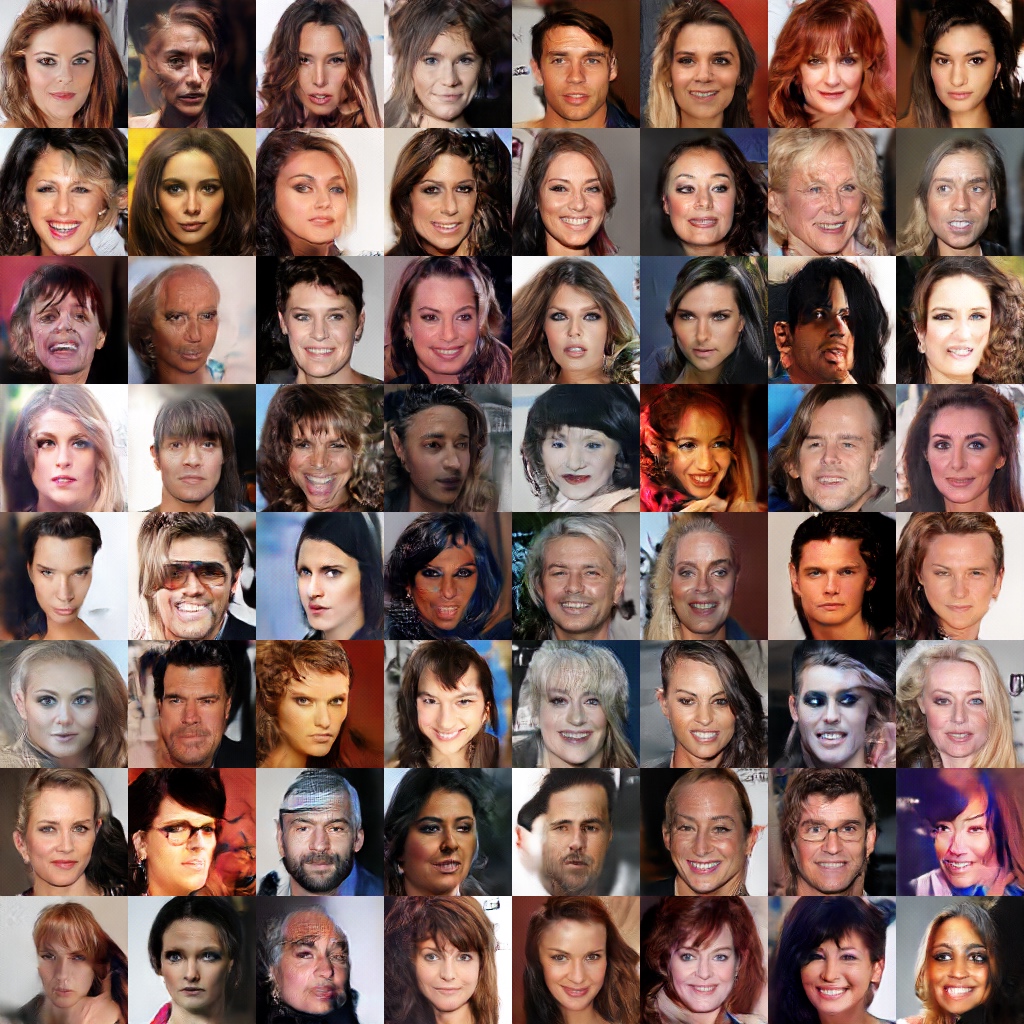}
		\caption{FID = 45.2} 
	\end{subfigure}
	\caption{Examples generated by GANs on \textsc{celeba-hq-128} dataset.}
	\label{fig:examples_celeba_fid}
\end{figure}

\begin{figure}[h]
	\centering
	\begin{subfigure}{0.32\textwidth}
		\includegraphics[width=\textwidth]{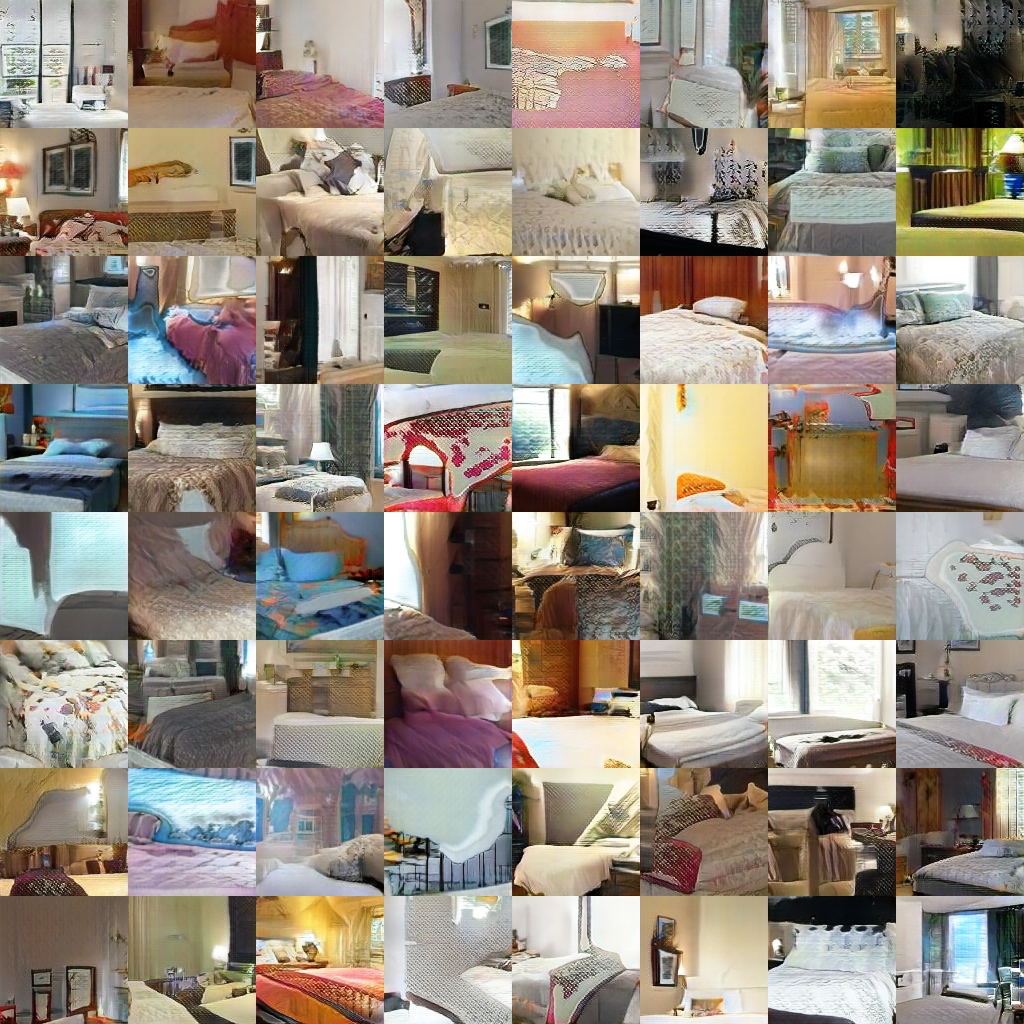}
		\caption{FID = 40.4} 
	\end{subfigure}
	\begin{subfigure}{0.32\textwidth}
		\includegraphics[width=\textwidth]{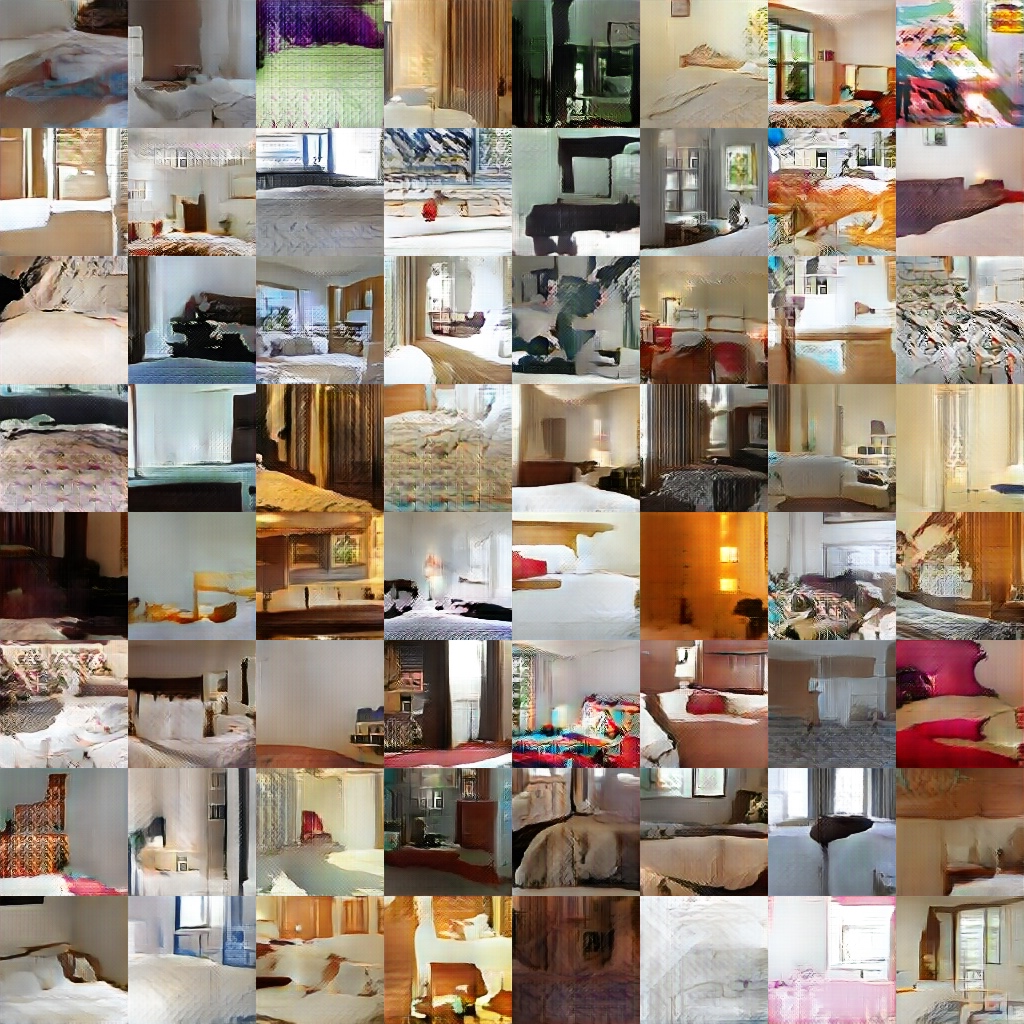}
		\caption{FID = 60.7} 
	\end{subfigure}
	\begin{subfigure}{0.32\textwidth}
		\includegraphics[width=\textwidth]{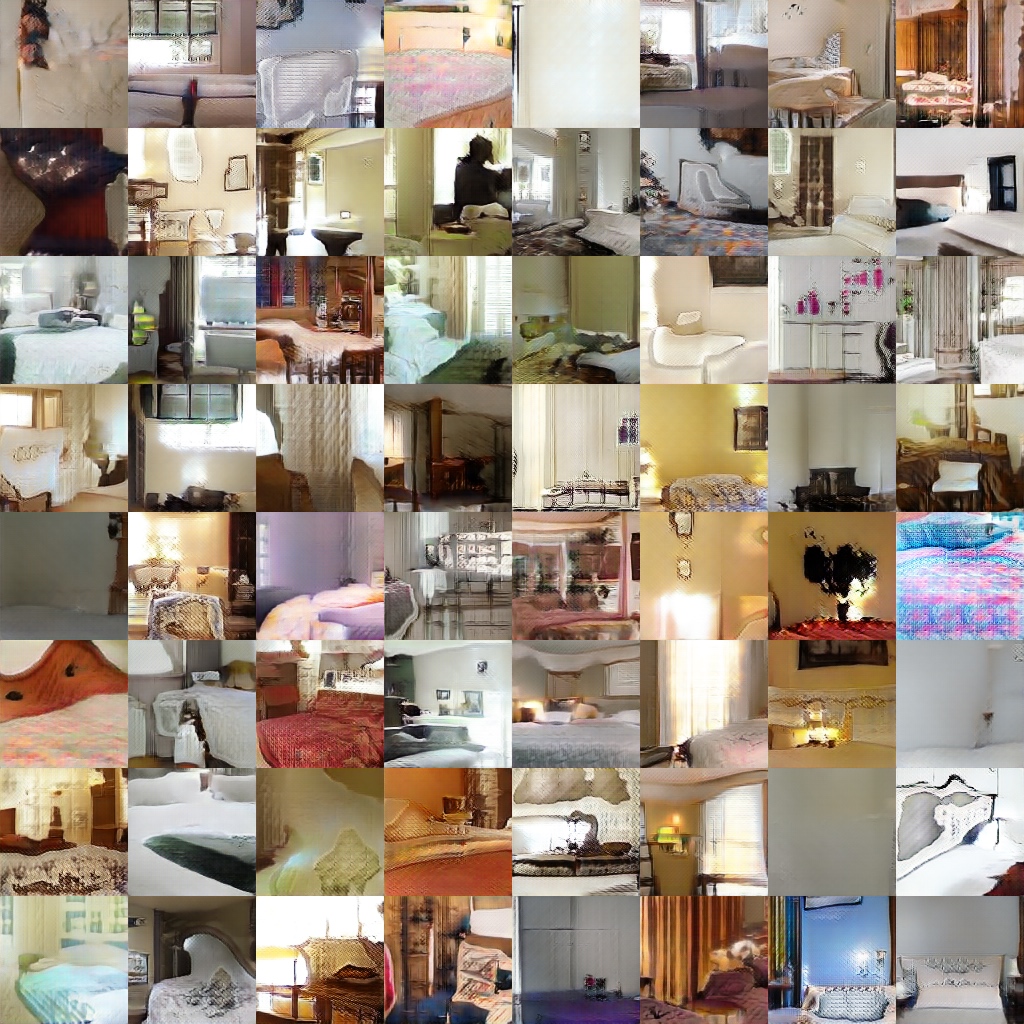}
		\caption{FID = 80.2} 
	\end{subfigure}
	\caption{Examples generated by GANs on $\lsun$ dataset.}
	\label{fig:examples_lsun_fid}
\end{figure}

\begin{figure}[h]
	\centering
	\begin{subfigure}{0.32\textwidth}
		\includegraphics[width=\textwidth]{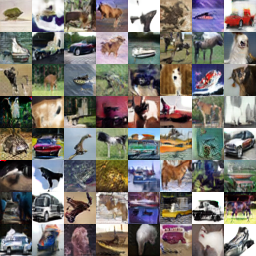}
		\caption{FID = 22.7} 
	\end{subfigure}
	\begin{subfigure}{0.32\textwidth}
		\includegraphics[width=\textwidth]{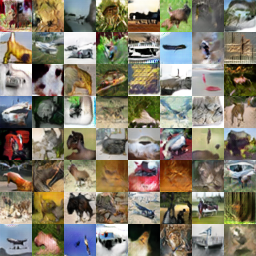}
		\caption{FID = 33.0} 
	\end{subfigure}
	\begin{subfigure}{0.32\textwidth}
		\includegraphics[width=\textwidth]{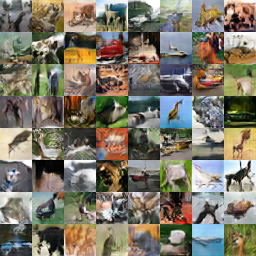}
		\caption{FID = 42.6} 
	\end{subfigure}
	\caption{Examples generated by GANs on \textsc{cifar10} dataset.}
	\label{fig:examples_cifar_fid}
\end{figure}

To visualize the FID score on each dataset, Figure~\ref{fig:examples_celeba_fid}, Figure~\ref{fig:examples_lsun_fid} and Figure~\ref{fig:examples_cifar_fid} show the generated examples by GANs. We select the examples from the best FID run, and then increase the FID score for two more plots.

\begin{table}[h]
\centering
\caption{SNDCGAN parameters}
\begin{tabular}{lllllllll}
  \toprule
  \textsc{Dataset} & \textsc{Median} & \textsc{Best}& \textsc{LR$(\times 10^{-3})$} & \textsc{$\beta_1$} & \textsc{$\beta_2$} & \textsc{$n_{disc}$} & \textsc{$\lambda$} & \textsc{Norm} \\\toprule
  \textsc{cifar10} & 29.75 & 28.66 & 0.100 & 0.500 & 0.999 & 1 & - & - \\
  \textsc{cifar10} & 36.12 & 33.23 & 0.200 & 0.500 & 0.999 & 1 & - & - \\
  \textsc{celeba-hq-128} & 66.42 & 63.13 & 0.100 & 0.500 & 0.999 & 1 & - & - \\
  \textsc{celeba-hq-128} & 67.39 & 64.59 & 0.200 & 0.500 & 0.999 & 1 & - & - \\
  $\lsun$ & 180.36 & 160.12 & 0.200 & 0.500 & 0.999 & 1 & - & - \\
  $\lsun$ & 188.99 & 162.00 & 0.100 & 0.500 & 0.999 & 1 & - & - \\ \midrule
  \textsc{cifar10} & 26.66 & 25.27 & 0.200 & 0.500 & 0.999 & 1 & - & SN \\
  \textsc{cifar10} & 27.32 & 26.97 & 0.100 & 0.500 & 0.999 & 1 & - & SN \\
  \textsc{celeba-hq-128} & 31.14 & 29.05 & 0.200 & 0.500 & 0.999 & 1 & - & SN \\
  \textsc{celeba-hq-128} & 33.52 & 31.92 & 0.100 & 0.500 & 0.999 & 1 & - & SN \\
  $\lsun$ & 63.46 & 58.13 & 0.200 & 0.500 & 0.999 & 1 & - & SN \\
  $\lsun$ & 74.66 & 59.94 & 1.000 & 0.500 & 0.999 & 1 & - & SN \\ \midrule
  \textsc{cifar10} & 26.23 & 26.01 & 0.200 & 0.500 & 0.999 & 1 & 1 & SN+GP \\
  \textsc{cifar10} & 26.66 & 25.27 & 0.200 & 0.500 & 0.999 & 1 & - & SN \\
  \textsc{celeba-hq-128} & 31.13 & 30.80 & 0.100 & 0.500 & 0.999 & 1 & 10 & GP \\
  \textsc{celeba-hq-128} & 31.14 & 29.05 & 0.200 & 0.500 & 0.999 & 1 & - & SN \\
  $\lsun$ & 63.46 & 58.13 & 0.200 & 0.500 & 0.999 & 1 & - & SN \\
  $\lsun$ & 66.58 & 65.75 & 0.200 & 0.500 & 0.999 & 1 & 10 & GP \\ \bottomrule
\end{tabular}
\label{tab:sndcgan_parameters}
\end{table}

\begin{table}[h]
\centering
\caption{ResNet19 parameters}
\begin{tabular}{lllllllll}
  \toprule
  \textsc{Dataset} & \textsc{Median} & \textsc{Best}& \textsc{LR$(\times 10^{-3})$} & \textsc{$\beta_1$} & \textsc{$\beta_2$} & \textsc{$n_{disc}$} & \textsc{$\lambda$} & \textsc{Norm} \\\toprule
  \textsc{celeba-hq-128} & 43.73 & 39.10 & 0.100 & 0.500 & 0.999 & 5 & - & - \\
  \textsc{celeba-hq-128} & 43.77 & 39.60 & 0.100 & 0.500 & 0.999 & 1 & - & - \\
  $\lsun$ & 160.97 & 119.58 & 0.100 & 0.500 & 0.900 & 5 & - & - \\
  $\lsun$ & 161.70 & 125.55 & 0.100 & 0.500 & 0.900 & 5 & - & - \\ \midrule
  \textsc{celeba-hq-128} & 32.46 & 28.52 & 0.100 & 0.500 & 0.999 & 1 & - & LN \\
  \textsc{celeba-hq-128} & 40.58 & 36.37 & 0.200 & 0.500 & 0.900 & 1 & - & LN \\
  $\lsun$ & 70.30 & 48.88 & 1.000 & 0.500 & 0.999 & 1 & - & SN \\
  $\lsun$ & 73.84 & 60.54 & 0.100 & 0.500 & 0.900 & 5 & - & SN \\ \midrule
  \textsc{celeba-hq-128} & 29.13 & - & 0.100 & 0.500 & 0.900 & 5 & 1 & LN+DR \\
  \textsc{celeba-hq-128} & 29.65 & - & 0.200 & 0.500 & 0.900 & 5 & 1 & GP \\
  $\lsun$ & 55.72 & - & 0.200 & 0.500 & 0.900 & 5 & 1 & LN+GP \\
  $\lsun$ & 57.81 & - & 0.100 & 0.500 & 0.999 & 1 & 10 & SN+GP \\ \bottomrule
\end{tabular}
\label{tab:resnet19_parameters}
\end{table}

\begin{table}[h]
\centering
\caption{ResNet CIFAR parameters}
\begin{tabular}{lllllllll}
  \toprule
  \textsc{Dataset} & \textsc{Median} & \textsc{Best} & \textsc{LR$(\times 10^{-3})$} & \textsc{$\beta_1$} & \textsc{$\beta_2$} & \textsc{$n_{disc}$} & \textsc{$\lambda$} & \textsc{Norm} \\\toprule
  \textsc{cifar10} & 31.40 & 28.12 & 0.200 & 0.500 & 0.999 & 5 & - & - \\
  \textsc{cifar10} & 33.79 & 30.08 & 0.100 & 0.500 & 0.999 & 5 & - & - \\ \midrule
  \textsc{cifar10} & 23.57 & 22.91 & 0.200 & 0.500 & 0.999 & 5 & - & SN \\
  \textsc{cifar10} & 25.50 & 24.21 & 0.100 & 0.500 & 0.999 & 5 & - & SN \\ \midrule
  \textsc{cifar10} & 22.98 & 22.73 & 0.200 & 0.500 & 0.999 & 1 & 1 & SN+GP \\
  \textsc{cifar10} & 23.57 & 22.91 & 0.200 & 0.500 & 0.999 & 5 & - & SN \\ \bottomrule
\end{tabular}
\label{tab:resnet_cifar_parameters}
\end{table}

\section{Relative Importance of Optimization Hyperparameters}
\label{sec:parameter-heat-plot-appendix}
\begin{figure}[h]
	\centering
	\begin{subfigure}{1.0\textwidth}
		\includegraphics[width=\textwidth]{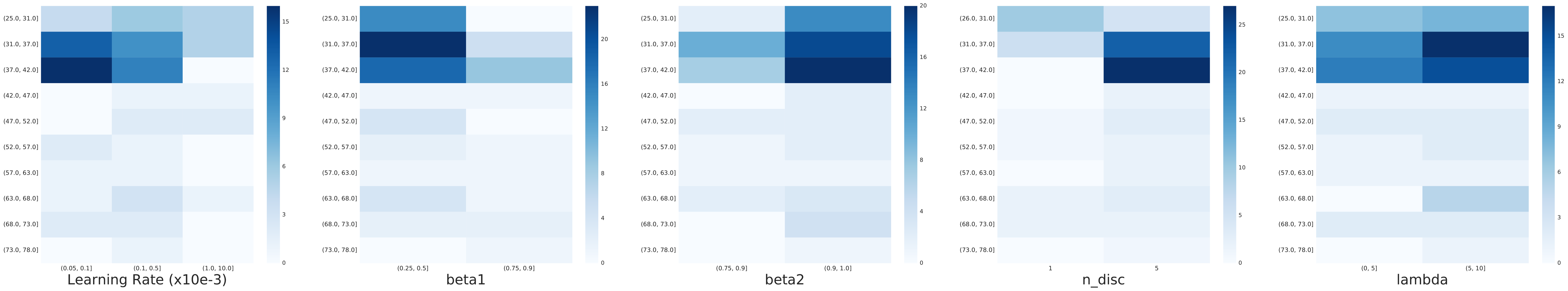}
		\caption{FID score of SNDCGAN on \textsc{cifar10}} 
	\end{subfigure}
	\begin{subfigure}{1.0\textwidth}
		\includegraphics[width=\textwidth]{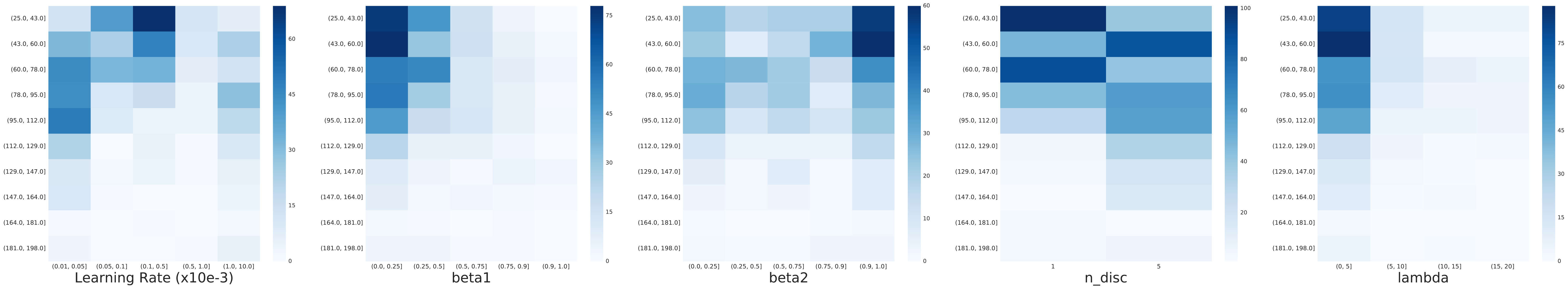}
		\caption{FID score of SNDCGAN on \textsc{celeba-hq-128}} 
	\end{subfigure}
	\begin{subfigure}{1.0\textwidth}
		\includegraphics[width=\textwidth]{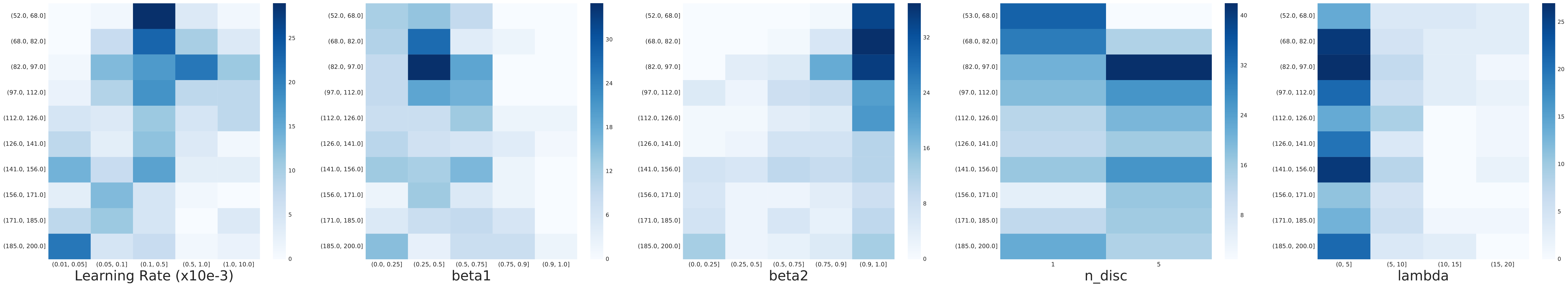}
		\caption{FID score of SNDCGAN on $\lsun$} 
	\end{subfigure}
	\begin{subfigure}{1.0\textwidth}
		\includegraphics[width=\textwidth]{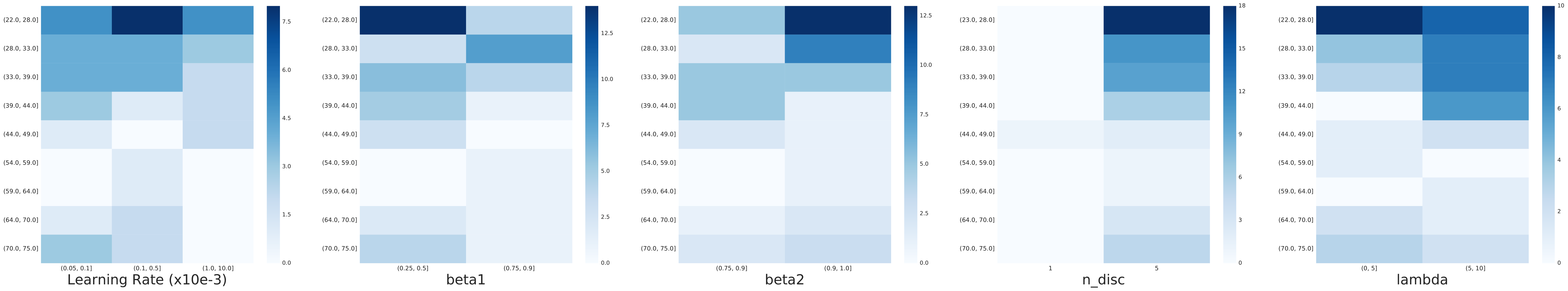}
		\caption{FID score of ResNet CIFAR on \textsc{cifar10}} 
	\end{subfigure}
	\begin{subfigure}{1.0\textwidth}
		\includegraphics[width=\textwidth]{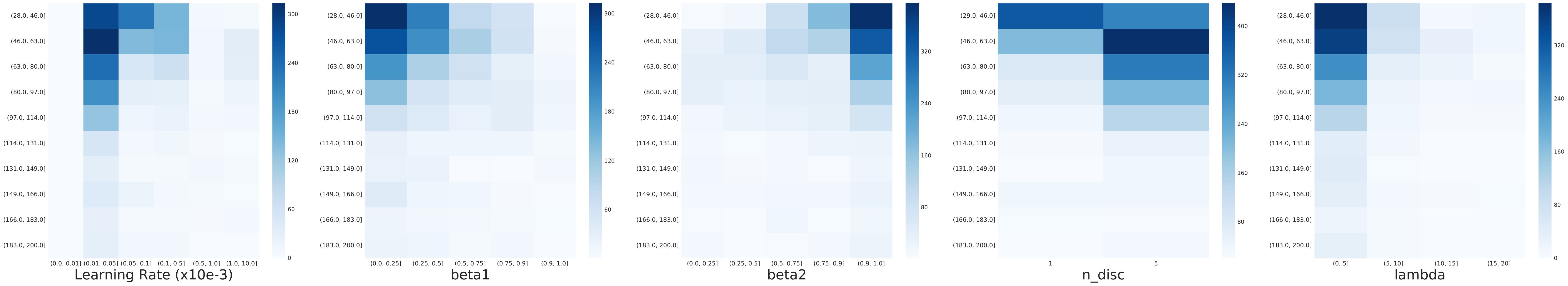}
		\caption{FID score of ResNet19 on \textsc{celeba-hq-128}} 
	\end{subfigure}
	\begin{subfigure}{1.0\textwidth}
		\includegraphics[width=\textwidth]{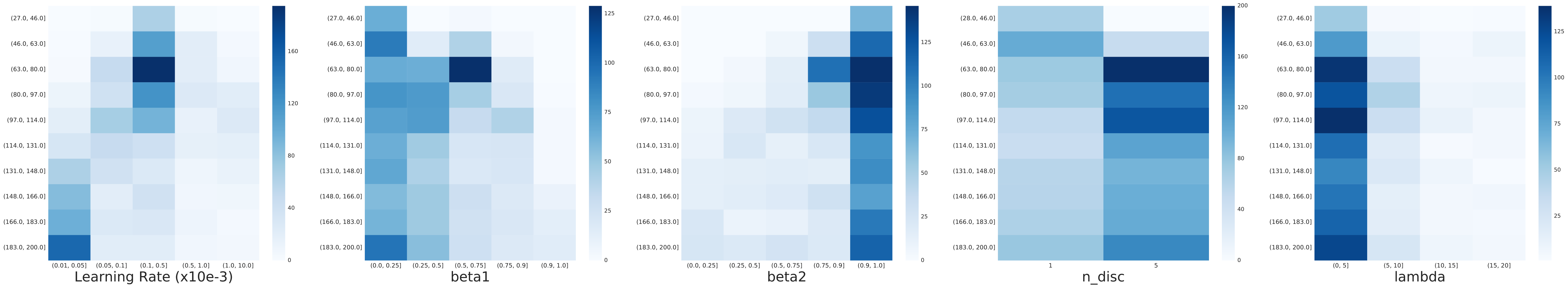}
		\caption{FID score of ResNet19 on $\lsun$} 
	\end{subfigure}
	\caption{Heat plots for hyper-parameters on each architecture and dataset
  combination.}
	\label{fig:heat_plot_param_fid}
\end{figure}

For each architecture and hyper-parameter we estimate
its impact on the final FID.
Figure~\ref{fig:heat_plot_param_fid} presents heatmaps for hyperparameters, namely the learning rate,
$\beta_1$, $\beta_2$, $n_{disc}$, and $\lambda$ for each combination
of neural architecture and dataset.

\end{document}